\documentclass[11pt]{article}
\usepackage{eamt26}
\usepackage{times}
\usepackage{url}
\usepackage{latexsym}
\usepackage[small,bf]{caption} 
\usepackage{graphicx}

\usepackage{booktabs}
\usepackage{makecell}
\usepackage{tabularx}
\usepackage{inconsolata}
\usepackage{microtype}
\usepackage{hyperref}
\usepackage{xcolor}
\definecolor{darkblue}{rgb}{0, 0, 0.5}
\hypersetup{
    colorlinks=true,
    citecolor=darkblue,
    linkcolor=darkblue,
    urlcolor=darkblue
} 
\usepackage{xcolor}
\usepackage{comment}
\usepackage{enumitem}
\usepackage{multirow}
\usepackage{fontawesome5}
\usepackage{ragged2e}
\usepackage{todonotes}
\usepackage{tcolorbox}
\tcbuselibrary{skins}
\usepackage{soul}
\usepackage{xcolor}

\makeatletter
\def\@biblabel#1{\hspace{-\leftmargin}}
\makeatother

\newcommand{\fem}[1]{%
  \tcbox[on line, arc=3pt, colback=pink!35, colframe=pink!35,
         boxsep=0pt, left=1pt, right=1pt, top=0pt, bottom=0pt]{#1}}
\newcommand{\masc}[1]{%
  \tcbox[on line, arc=3pt, colback=cyan!20, colframe=cyan!20,
         boxsep=0pt, left=1pt, right=1pt, top=0pt, bottom=0pt]{#1}}

\title{Explaining \textsc{GAND}: A Resource on Gender-Ambiguous Natural Data \\ \& Contrastive Attribution}

\author{Jani\c ca Hackenbuchner, Jasper Degraeuwe, Arda Tezcan and Joke Daems\\
Language and Translation Technology Team (LT$^3$)\\
Department of Translation, Interpreting and Communication\\
Ghent University, Belgium\\
{\tt \{firstname.lastname\}@ugent.be}
}

\begin{document}
\maketitle
\begin{abstract}

Machine translation (MT) systems 
continue to produce gender-biased translations. In a time where self-expression is paramount, mistranslations based on default behaviour and stereotyping can lead to harm for users of these systems. To better understand how these systems translate gender in the absence of clear gender cues, we need benchmarking resources that reflect gender-ambiguous scenarios in a natural way. To this end, we present \textsc{GAND}, a gender-ambiguous natural data benchmarking resource for MT consisting of English source sentences, specifically designed to analyse the influence of contextual cues on gender in translation. We leverage \textsc{GAND} to conduct an interpretability analysis:
we translate a subset of \textsc{GAND} into two grammatical gender languages and extend these with manually crafted contrastive translations. A following feature attribution analysis reveals source words in context that inform the gender translation of an ambiguous referent entity in the target translation.


\end{abstract}

\section{Introduction}

Machine Translation (MT) systems and Large Language Models (LLMs) have long been shown to exhibit gender-biased behaviour, typically defaulting to generic masculine forms or making assumptions based on gender stereotypes ~\cite{blodgett-etal-2020,saunders-byrne-2020,savoldi-etal-2021,kotek-etal-2023,vanmassenhove-2024,gkovedarou_gender_2025}. When translating from notional languages with limited gender marking (e.g., English) to grammatical languages with overt gender distinctions (e.g., German or Spanish), MT systems tend to assign grammatical gender to ambiguous referent entities. This is a phenomenon that Rarrick et al. ~\shortcite[p.~845]{rarrick-etal-2023} dub ``Arbitrarily Gender-Marked Entities'', where referent (or animate) entities are arbitrarily gender marked, even though it is ``not implied in the source''. In such absence of clear grammatical gender cues 
in an ambiguous source, MT systems overwhelmingly default to generic masculine forms or stereotype-based translations ~\cite{hackenbuchner-etal-2025-genderous,mastromichalakis-etal-2025}, whereas humans have been shown to interpret ambiguous referents more diversely ~\cite{hackenbuchner-etal-2025-clin}.

\begin{figure}[t]
  \includegraphics[width=\columnwidth]{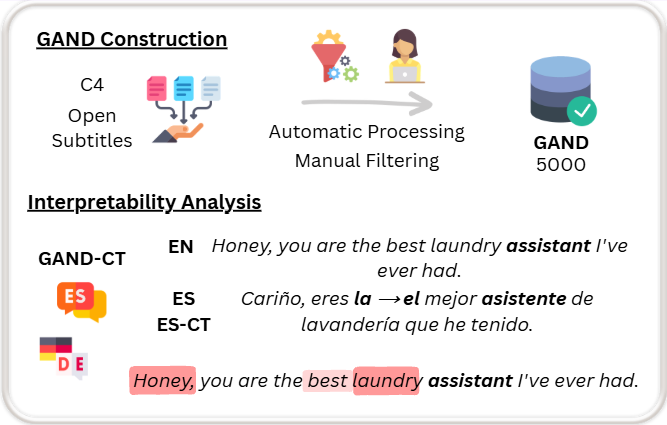}
  \caption{Infographic of the creation of GAND and a simplified visualisation of the interpretability analysis via contrastive translations.}
  \label{fig:1}
\end{figure}

The last decade has shown extensive research on gender bias in MT, with numerous data resources published to benchmark state-of-the-art MT systems and general-purpose foundation models to assess and mitigate gender bias ~\cite{savoldi-etal-2025-decade}, a selection of which is presented in Section~\ref{benchmarks}. However, among this vast amount of data resources, there continues to be a lack of datasets that capture the fully gender-ambiguous phenomenon in a natural setting, as these kinds of data cannot be easily collected in an automated way. The phenomenon therefore continues to be underexplored, despite being crucial from a translation perspective. As ambiguous sentences are very common in translation and have been found to be particularly challenging for MT systems \cite{saunders-olsen-2023}, we need suitable data resources to study them.

Alongside the need for suitable data resources, there is an increasing demand for explainable AI (XAI) ~\cite{räuker2023,ferrando2024} to understand when a model is making a biased prediction based on sensitive attributes and, with that, inform social bias mitigation strategies during model training. In an attempt to better understand and reveal mechanisms underlying gendered choices in MT systems, gender bias in MT (and recently also speech translation; Conti et al. ~\shortcite{conti-etal-2025-unheard}) is increasingly being examined by means of interpretability techniques. Among these techniques, coreference resolution has been studied by means of attention mechanisms ~\cite{costa-jussa-etal-2022,manna-etal-2025-paying} as well as feature attribution ~\cite{sarti-etal-2023-inseq,attanasio-etal-2023-tale}.

In this paper, we address these two demands and present \textsc{GAND}\footnote{The resource, programming scripts used for its compilation, and scripts used for the analyses presented in this paper are made available in a GitHub repository at \url{https://github.com/jhacken/GAND}.}\footnote{The dataset is available on HuggingFace: \url{https://huggingface.co/datasets/jhacken/GAND}.} 
a data resource on \textbf{G}ender-\textbf{A}mbiguous \textbf{N}atural \textbf{D}ata, to analyse the influence of contextual cues on gender in translation.
\textsc{GAND} consists of 5047 English sentences that have been compiled from natural data and are strictly gender-ambiguous for a given singular referent. 
Our methodology combines automated data filtering and rule-based cleaning with manual verification, showing the difficulty of compiling a resource of strictly gender-ambiguous but linguistically rich data. The significant contribution of this dataset is the shift from (primarily) synthetic templates to natural data, which can be used to assess inherent model bias in a realistic setting. This resource forms the basis of a larger research project, which aims to automatically identify the elements of a source sentence that most strongly influence MT gender inflections for a specific referent ~\cite{hackenbuchner-etal-2024-project}. The project is diagnostic rather than normative: in ambiguous scenarios, there is not one `right' translation, rather, users should be made aware of scenarios where ambiguity in the source is more likely to lead to explicitly gendered outputs.

Considering that context provides relevant information in the process of translation, we want to find salient context that affects the translation in terms of gender to better understand this influence. To this end, we leverage \textsc{GAND} to assess whether explanations can be used to detect when a model is making a gender-biased prediction based on salient attributions. For this, we translate a subset of 1000 sentences into two grammatical gender languages, Spanish and German, and manually extend these with contrastive translations ~\cite{yin-neubig-2022-interpreting}. On these, we conduct an interpretability-informed analysis to identify which contextual cues in the source influence the gender translation in the target. The data generation process for the \textsc{GAND} dataset, along with the interpretability analysis conducted in this study, is illustrated in Figure~\ref{fig:1}.

Our contribution in this paper, therefore, is two-fold: (i) we present \textsc{GAND}, \textbf{a first-of-its-kind extensive data resource on gender-ambiguous natural data}, suitable for benchmarking MT, and (ii) we conduct an \textbf{interpretability-informed analysis} by means of contrastive translations and saliency attribution to \textbf{showcase contextual cues that influence the gender in translation}.


\section{Related Research}\label{related}

\subsection{Existing Benchmarks}\label{benchmarks}
The following contains a brief discussion of key benchmarks, yet does not claim to be exhaustive. They are marked in the text with \textsuperscript{*} if based on artificial data, and \textsuperscript{°} if based on natural data. An example sentence showcasing the style of each of the data resources presented here is provided in Appendix \ref{app:benchmark_examples} Table \ref{app:tab:dataset_examples}. The majority of existing data resources consist of gender-unambiguous sentences (w.r.t. a specific referent entity), where coreference resolution is examined, including WinoBias\textsuperscript{*} ~\cite{zhao-etal-2018-gender}, WinoMT\textsuperscript{*} ~\cite{stanovsky-etal-2019}, SimpleGen\textsuperscript{*} ~\cite{renduchintala-etal-2021}, BUG\textsuperscript{°} (based on Wikipedia and medical data; Levy et al. ~\shortcite{levy-etal-2021}), MT-GenEval\textsuperscript{°} (based on Wikipedia; Currey et al. ~\shortcite{currey-etal-2022}), or MiTTens\textsuperscript{*°} ~\cite{robinson-etal-2024}, \textit{inter alia}.

Recent benchmarks contain both unambiguous and ambiguous scenarios, such as GLITTER\textsuperscript{°} (based on Wikipedia and plural referents; Pranav et al. ~\shortcite{pranav-etal-2025}) and GeNTE\textsuperscript{°} and mGeNTE\textsuperscript{°} (based on EuroParl and including both singular and plural referents; Piergentili et al.~\shortcite{piergentili-etal-2023-hi} and Savoldi et al. ~\shortcite{savoldi-etal-2025-mgente}). First-person singular gender ambiguity w.r.t. a speaker (e.g., ``\textit{I} was born...''), and unambiguous otherwise, can be evaluated with the Must-SHE corpus\textsuperscript{°} (based on TedTalks; Bentivogli et al. ~\shortcite{bentivogli-etal-2020}) and the Arabic parallel gender corpus\textsuperscript{°} (based on OpenSubtitles; Habash et al. ~\shortcite{habash-etal-2019}). Gender-ambiguity can further be evaluated in GATE\textsuperscript{°*} ~\cite{rarrick-etal-2023}, a linguistically-constructed challenge set that contains both singular and plural referents as well as speaker-ambiguity, in GENDEROUS\textsuperscript{*} ~\cite{hackenbuchner-etal-2025-genderous}, a small, strictly artificial challenge set, or in GAMBIT\textsuperscript{*} ~\cite{mastromichalakis-etal-2025} and GAMBIT+\textsuperscript{*} ~\cite{filandrianos-etal-2025}, both LLM-generated and manually vetted datasets. To highlight certain differences of GAND in relation to existing datasets (e.g., GATE), we provide specific examples in Appendix \ref{app:benchmark_examples}.

Mastromichalakis~\shortcite[p.~32227]{mastromichalakis-etal-2025} initially intended to generate a realistic, natural gender-ambiguous dataset but concluded that ``artificial generation [was] the only viable approach for ensuring both full occupational coverage and a variety of textual styles'', and is certainly a less time-consuming approach. Rarrick et al. ~\shortcite[p.~848]{rarrick-etal-2023} similarly voice the difficulty of this while filtering for ambiguous natural data for GATE, as their success was ``depending on the relative ease of finding such sentences'' and that they resorted to manually ``modif[ying] [sentences] slightly to fit the requirements''.

To date, therefore, there is no extensive dataset that captures the fully gender-ambiguous phenomenon w.r.t. a singular referent entity in a realistic, natural setting. 

\begingroup
\renewcommand{\arraystretch}{1.25} 
\begin{table*}[t]
  \centering
  \begin{tabularx}{\textwidth}{X}
  \toprule
    \textbf{GAND Resource} \\
    \midrule
    ``A report from the country's correctional \textit{investigator} says overcrowding is one factor in why Canada's prisons are becoming more violent.'' \\
    ``The \textit{clerk} gave the suspect an undisclosed amount of money.''\\
     ``His plan allowed for an in-home \textit{helper} who came every day to get Agnes up and be a personal assistant and to assist Agnes in performing her activities of daily living–grooming, toileting, dressing, eating, walking and whatever else was needed.''\\
   \bottomrule
  \end{tabularx}
  \caption{\label{GAND-examples}
    Three example sentences included in the \textsc{GAND} dataset. The (ambiguous) referent is marked in italics.
  }
\end{table*}
\endgroup

\subsection{Interpretability-informed Research}\label{interpretability_research}
\paragraph{Anaphora Resolution}
Research on the majority of the above-presented benchmarks has predominantly assessed to what extent MT systems and LLMs adhere to coreference chains, while more recent benchmarks further focus on gender-neutral and gender-fair translation generation. Coreference resolution in the WinoMT corpus ~\cite{stanovsky-etal-2019} has been informed by feature attribution in Attanasio et al. ~\shortcite{attanasio-etal-2023-tale} and by means of attention in Manna et al. ~\shortcite{manna-etal-2025-paying}. Sarti et al. ~\shortcite{sarti-etal-2023-inseq} and Wisniewski et al. ~\shortcite{wisniewski-etal-2022} conducted an interpretability-informed analysis on their own handcrafted test sets. Results reveal accurate gender disambiguation to critically depend on correct coreference resolution ~\cite{wisniewski-etal-2022,manna-etal-2025-paying}. When models fail to correctly resolve coreference, they lead to incorrect predictions by defaulting to masculine translations ~\cite{sarti-etal-2023-inseq,attanasio-etal-2023-tale}.

While contextual cues in a source have been shown to influence the target gender an MT system assigns in a target language, this can be explained by means of coreference ~\cite{kocmi-etal-2020} but not yet in ambiguous scenarios.

\paragraph{Contrastive Explanations}
Contrastive explanations, as an interpretability methodology, have been shown to outperform others, demonstrating which input tokens lead a model to predict one output (\textit{target}) instead of another (\textit{foil}) ~\cite{yin-neubig-2022-interpreting}. Contrastive explanations have successfully been studied for gender (e.g., \textit{Why is M predicted instead of F?}) in previous translation research ~\cite{vamvas-sennrich-2021,sarti-etal-2023-inseq,conti-etal-2025,hackenbuchner2025-arxiv}. Based on the Must-SHE corpus ~\cite{bentivogli-etal-2020}, Conti et al. ~\shortcite{conti-etal-2025-unheard,conti-etal-2025} leverage contrastive feature attribution in speech translation to reveal that, to accurately translate gender, the model relies on first-person pronouns to link gendered terms back to the speaker. Based on a small sample of gender-ambiguous sentences, Hackenbuchner et al. ~\shortcite{hackenbuchner2025-arxiv} leverage contrastive translations to study plausibility, or `human-interpretability' ~\cite{lage-etal-2019}, which refers to the alignment between model rationales and salient source words identified by human annotators. They show that the model and humans largely align, being influenced by similar content words in (an ambiguous) context. In this paper, we build on previous research, leveraging \textsc{GAND} to conduct an interpretability-informed analysis by means of contrastive translations. In the absence of clear gender and/or coreferent cues, we find salient source words in the sentence context that inform the model's decision to opt for a specific gender in translation instead of another.

\section{Data Compilation}\label{data}

To date, there is no dataset that captures the \textit{fully gender-ambiguous phenomenon w.r.t. a singular referent entity in a realistic, natural setting}. To fill this gap, we present \textsc{GAND}, \textbf{a dataset compiling 5047 English natural gender-ambiguous sentences referring to a singular referent entity}. To cover a variety of topics and linguistic diversity, \textsc{GAND} has been compiled from two natural sources: C4 Common Crawl\footnote{\url{https://huggingface.co/datasets/allenai/c4}} (EN subset) and Open Subtitles\footnote{\url{https://opus.nlpl.eu/OpenSubtitles/corpus/version/OpenSubtitles} } (monolingual EN data) from the OPUS project\footnote{\url{https://opus.nlpl.eu/}}. Due to a lack of automation for processing data for gender-ambiguity, which has been unsuccessfully attempted in previous research due to its difficulty in nature, the data resource presented here has been meticulously compiled by a detailed automatic processing methodology, followed by rigorous manual vetting of thousands of sentences. The specific steps undertaken are outlined in more detail in the following subsections.

\subsection{Data Filtering}\label{data_filtering}

Firstly, to filter for singular referent entities, we manually compiled pre-defined lists, totaling to 183 singular referents, such as `journalist' (see Appendix \ref{GAND_rules} for a full list). Male- or female-associated referents (e.g., `architect' and `dancer', respectively) have been compiled from previous research on word embeddings ~\cite{Bolukbasi-etal-2016,stanovsky-etal-2019,caliskan-etal-2022}. To add to these and particularly to find referent entities without a specific binary gender association, we compiled a manual list (cross-referencing this with terms that were considered `neutral' by a state-of-the-art generative LLM\footnote{OpenAI's \href{https://chatgpt.com/}{ChatGPT-4}; state-of-the-art at the time of use.} and that, in previous research, have not been classified to have a binary gender association). 
This compilation includes `neutrally-associated' referents, such as `visitor' or `traveler'. This allows for a roughly equal split between singular referents with the final list consisting of words with male or female-associated embeddings (count of 72 (39.5\%) and count of 57 (31\%) respectively), or ones that can be considered neutral (count of 54, representing 29.5\%). 

In the first processing step, we load the datasets (C4 and OpenSubtitles) and iterate over the texts in the (randomly shuffled) dataset, extracting all sentences that (1) contain at least one of the referent words, (2) start with an uppercased letter, (3) do not contain a (semi-)colon, and (4) meet the minimum and maximum sentence length criteria (minimum of 5 tokens and a maximum length of 50 tokens).

\subsection{Automated Data Cleaning}
In the following processing step, we apply a set of handcrafted rules to exclude sentences that (1) are unsuitable (e.g., no conjugated verb or incorrect part of speech) or that (2) might include a (co)reference that provides cues that allow for disambiguation of the referent. To exclude these, we check whether any proper nouns or ``gender (pro)nouns'' (a manually-compiled list, provided in Appendix \ref{GAND_rules}) are in reference to the referent. These rules to exclude sentences with anaphora that disambiguate the referent were necessary because we particularly want to keep sentences with coreference towards \textit{other} referents in the sentence, as shown by the third sentence in Table \ref{GAND-examples},
meaning that we could not simply filter out personal gender pronouns (`he', `she', `his', `her', etc.). To perform the morphosyntactic analysis of the sentences, we use Stanza's NLP toolkit\footnote{\url{https://stanfordnlp.github.io/stanza/}} ~\cite{qi_stanza_2020}. A more detailed list of the ten coreference checks is provided in Appendix \ref{GAND_rules} and in the GitHub repository, where we additionally highlight examples and include a full list of sentences that were excluded following this automatic processing step. The aim of this resource compilation is not to collect \textit{every possible} gender-ambiguous instance within C4 and Open Subtitles but rather to create a resource vast enough for studying and benchmarking systems on the topic of gender ambiguity. In total, we processed 2.75 million sentences for OpenSubtitles and 20.4 million sentences for C4, of which only 0.3\% and 2.25\% of sentences, respectively, passed the data filtering and cleaning process. More detailed information about the filtering numbers can be found in Appendix \ref{app:GAND_stats}.

\subsection{Manual Verification}

\begingroup
\renewcommand{\arraystretch}{1} 
\begin{table}
\centering
\begin{tabular}{c|c|c|c|c}
 & \textbf{\#total} & $\overline{\textbf{len}}$ & \textbf{\#sents} & \textbf{Source} \\
 \midrule
\multirow{2}{*}{\textsc{GAND}} & \multirow{2}{*}{5047} & 17.98 & 2908 & C4\\
& & $\sigma$ 8.52 & 2139 & OpenSubs \\
\bottomrule
\end{tabular}
\\[0.5em]
\begin{tabular}{cc|c|c|c}
\multicolumn{2}{c|}{Ref embedding} & \faMale & \faFemale & \faUser \\
\midrule
& Counts & 2222 & 1601 & 1224 \\
\bottomrule
\end{tabular}
\caption{\label{tab:GAND-stats}
 \textsc{GAND} dataset statistics: total number of sentences (\#total), average length per sentence ($\overline{len}$), standard deviation ($\sigma$), and the number of sentences (\#sents) per source corpus. The bottom half represents how many sentences in \textsc{GAND} include a referent from the masculine, feminine, or neutral word embedding/gender list.
  }
\end{table}
\endgroup

To ensure high quality, in the final step, a subset of 14,511 sentences that passed the automatic processing step was manually checked. Even though the designed automatic processing steps proved to be relatively successful at filtering out hundreds of thousands of sentences, they are not exhaustive. 
The manual check was conducted by the main author at a rate of $\sim$200 sentences per hour. In total, 34.8\% passed this verification step, leading to the 5047 sentences included in \textsc{GAND}, as exemplified in Table \ref{GAND-examples}. 
This manual labour has been purposefully conducted to ensure a high-quality dataset that can be presented as a resource to the community. Manually excluded sentences include sentences with errors, anaphora disambiguating the referent, referent names, or with indirect references (where the referent is not being referred to directly, and could thus be replaced by anyone or by a plural term, e.g. `a user'). An example list of sentences that were manually excluded can be found in Appendix \ref{app:exclusions} Table \ref{app:tab:exclusions}. Table \ref{tab:GAND-stats} summarises the main dataset statistics. The slight mismatch in referent gender association is carried on to the source sentences. Nevertheless, we decided to keep all manually vetted instances to include as many as possible.  
A detailed list of the number of sentences per referent entity found in \textsc{GAND} is provided in the GitHub repository.

\section{Interpretability Analysis}\label{interpretability}

To find contextual cues that influence the gender in translation, 
we leverage a randomly selected subset of 1000 of \textsc{GAND}'s sentences\footnote{In this subset, 442 sentences have a male-associated, 330 a female-associated, and 292 a neutral referent.}, which we dub \textsc{GAND-CT}, to conduct an interpretability-informed analysis by means of contrastive translations, following previous work ~\cite{conti-etal-2025-unheard,hackenbuchner2025-arxiv}. We investigate salient source words in the sentence context that inform the model's decision to opt for a specific gender (\textit{target}) in translation instead of another (\textit{foil}). We analyse whether saliency-informed explanations can be used to reliably detect when a model is making a gender-biased prediction for our dataset.

\subsection{Contrastive Translation Creation}\label{CTs}
\paragraph{Translation Pipeline} We translate \textsc{GAND-CT} into two grammatical gender target languages, German (DE) and Spanish (ES), using the OPUS-MT, transformer-based neural MT models\footnote{\url{https://huggingface.co/Helsinki-NLP/opus-mt-en-de}; 

\url{https://huggingface.co/Helsinki-NLP/opus-mt-en-es}} ~\cite{tiedemann2020}, leading to 2000 target translations. Due to this methodology requiring the examination of internal model behaviour, it limits the choice of transformer-based NMT models available for analysis, excluding recent state-of-the-art models, as well as limiting comparative analysis on decoder-only models.

\paragraph{Gender Annotation} The gender annotation of the referent word in the target translations was done manually 
according to the gender marked in the languages' leading dictionaries: Duden\footnote{Der Duden: \url{https://www.duden.de/}} for German and DLE\footnote{Diccionario de la lengua española: \url{https://dle.rae.es/}} for Spanish. Target referent gender was marked as either `masculine', `feminine', or `neutral'. 

\paragraph{Contrastive Translations} Starting with the EN $\rightarrow$ DE/ES translations, contrastive translations in terms of gender were manually created by the main author, who has (near-)native competence in the target languages and holds an MA degree in Specialised Translation. As the task is purely grammatical rather than subjective, only one linguist (with the required level of linguistic competence in the target languages) was required. Due to the naturalistic style of these sentences, more than one term must frequently be flipped to fully contrast the gender in translation. The creation of contrastive translations 
is exemplified for a simple scenario below (and in Figure \ref{fig:1}):

\begin{itemize}[topsep=3pt, itemsep=4pt, parsep=0pt, partopsep=3pt]
    \item EN Source: ``Honey, you are the best laundry \textit{assistant} I've ever had.''
    \item ES Translation: ``Cariño, eres \textit{la} mejor asistente de lavandería que he tenido.''
    \item ES Contrastive: ``Cariño, eres \textit{el} mejor asistente de lavandería que he tenido.''
\end{itemize}

Note, this is an example where the referent entity (\textit{asistente}) has a `dual gender' (traditionally both m/f), equally used for all genders and therefore annotated as `neutral', and is only gendered in conjunction with an e.g. article (\textit{la/el}). Target translations that were originally feminine, as shown in the example, were thus contrasted as masculine. In turn, originally masculine translations (e.g., \textit{médico}) were contrasted as feminine (\textit{médica}). If a target translation was originally neutral w.r.t. the referent (e.g., \textit{``¿Cómo se compara la escritura de thriller con su trabajo como }periodista\textit{?''}), these sentences were not contrasted and thus excluded from the attribution analysis (there was no contrastive gender creation as there is no (mis)gendered `error detection' necessary here). All sentences, translations, and annotations are, however, included in our data available in our GitHub repository for continued research.

Our further attribution analysis is thus based on 1000 source sentences and their 4000 target/contrastive sentences for DE and ES combined (excluding 41 `neutral' sentences for DE and 119 for ES).

\subsection{Saliency Attibution}

Using inseq\footnote{\url{https://github.com/inseq-team/inseq}} \texttt{version 0.7.0.dev0} ~\cite{sarti-etal-2023-inseq}, we compute saliency attribution of contrastive translations by contrasting: (i) the original MT translation (\textit{target}) and (ii) a translation contrasting in terms of gender (\textit{foil}). We analyse the contrastive probability difference of how much more likely it is for an original referent to be translated into one gender instead of the other, and we specifically focus on which source input tokens affect this probability. 
The computation of the contrastive gradient norm is based on work by Yin and Neubig ~\shortcite{yin-neubig-2022-interpreting}:

{
\abovedisplayskip=3pt
\belowdisplayskip=5pt
\[ g^c(x_i) = \Delta_{x_i} (q(y_t|\textbf{x}) - q(y_f|\textbf{x})) \]
}


where \(\textbf{x}\) is the input sequence embedding, \(y_t\) is the next token in the input sequence, \(q(y_t(\textbf{x}))\) is the model output for the token \(y_t\) and \((q(y_f(\textbf{x}))\) is the model output for foil token \(y_f\) given the input \(\textbf{x}\). This calculation tells how much an input token \(x_i\) influences the model to \textit{increase} the probability of \(y_t\) while \textit{decreasing} the probability of \(y_f\). Saliency attribution for the contrastive translation is thus computed as:

{
\abovedisplayskip=3pt
\belowdisplayskip=5pt
\[ S^c_{GN}(x_i) = ||g^c(x_i)||_{L_2} \]
}

The default applied here takes the L2 norm to aggregate gradient vectors and takes the probability of the next word.

Given the nature of encoder-decoder models, source tokens as well as previously generated target tokens influence the translations of the following words. To exclude the impact of attributions of previously generated target tokens, we left-align tokens and analyse the first (contrastive) different token in the MT output, such as \textit{la} $\rightarrow$ \textit{el} in the example above. With this, we ensure that the previously generated tokens (e.g., \textit{Cariño, eres}) remain the same in both translations, yielding attribution scores of zero for the contrastive translations of these preceding words.

\paragraph{Pre-Processing and Attribution Selection}
As we focus on source text saliency, we perform L1 normalisation to sum up the attribution scores to 1. We conduct pre-processing steps, as done in Hackenbuchner et al. ~\shortcite{hackenbuchner2025-arxiv}, to remove the target referent token (e.g., `assistant') to focus only on contextual cues, as well as end-of-sentence tokens, punctuation marks and a limited set of stopwords, including articles and determiners that do not contain any gender information (\textit{``}, \textit{''}, \textit{a}, \textit{an}, \textit{the}, \textit{this}, \textit{that}, \textit{these}, \textit{those}), as we are interested in \textit{content} words. We then merge sub-word tokens and add their individual attribution scores, as done in Manna et al. ~\shortcite{manna-etal-2025-paying}. We first calculate the total attribution score for a specific sentence
and then identify the minimum subset of input words whose cumulative attribution scores reach at least 15\% of the total attribution scores combined. This threshold is based on findings in Hackenbuchner et al. ~\shortcite{hackenbuchner2025-arxiv}, which, in a detailed comparison of different approaches, yielded this threshold to rank highest in terms of plausibility, where the model's choice of cues overlaps most with human-annotated perceptions of gender. We base our analysis on this threshold as we continuously aim to understand model choices in comparison to human perception with respect to gender in translation. Future work could include analyses based on alternative thresholds, depending on the research questions.

\subsection{Linguistic Analysis}
To better understand which \textit{types of words} in context influence a model's translation decision in terms of gender, we conduct a linguistic analysis of salient words. We use spaCy\footnote{The spaCy toolkit (\url{https://spacy.io/}) performs POS tagging based on the \href{https://universaldependencies.org/}{Universal Dependencies} annotation scheme.} to analyse both POS labels of salient words and to compute the dependency distance between each salient word in context and the target referent (e.g., between \textit{you} and \textit{assistant}). This distance counts the number of syntactic links (edges) separating two nodes in a given dependency tree and allows us to assess how close salient words are to the target referent in question, grammatically.

\begin{table}
\centering
\begin{tabular}{ccc|cc}
\multicolumn{3}{c|}{\textbf{\makecell[c]{OPUS-MT \\ Translation}}} & \multicolumn{2}{c}{\textbf{\makecell[c]{Contrastive \\ Prob. Difference}}} \\
\midrule
&&\textbf{Count}&\textbf{Mean} & \textbf{Std.}\\
&M & 893 & .56 & .24\\EN$\rightarrow$DE  &F & 64 &.36 & .33\\&N & 41 &-&-\\
\midrule
&M &814& .50 & .25\\EN$\rightarrow$ES  &F &65&.30 & .32\\&N & 120&-&-\\
\bottomrule
\end{tabular}
\caption{\label{tab:prob_differences}
    A summary of target translation counts in terms of gender per language combination, as well as the contrastive probability difference for the binary genders.}
\end{table}

\subsection{Results}

\subsubsection{Contrastive Probability Difference}

\begin{figure}
  \includegraphics[width=\columnwidth]{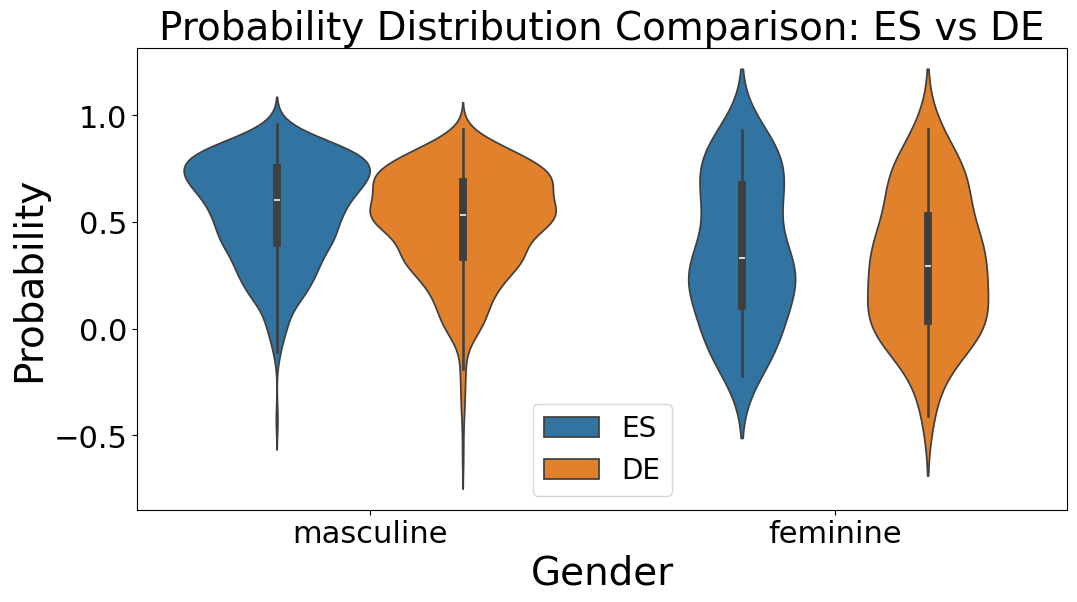}
  \caption{Contrastive probability difference of the model predicting one (gendered) token instead of the other.}
  \label{fig:violin_prob_difference}
\end{figure}

\paragraph{Overall Contrastive Probability Difference} For each contrastive translation of \textsc{GAND-CT} EN $\rightarrow$ DE/ES, we tally the \textit{contrastive probability difference} (CPD), which shows how much more likely the prediction of the original translated gender is in comparison to the contrastive gender. On average, the model shows \textbf{higher contrastive probability differences when translating a referent as masculine}, with a mean of 0.56 for EN-DE, meaning that the masculine translation of a referent entity is 56\% \textit{more probable} than the feminine one, and on average 50\% more probable for EN-ES. In comparison, the model shows, on average, lower CPDs when translating a referent as feminine, with a mean of being only 36\% more probable for a referent to be translated as feminine for EN-DE and 30\% for EN-ES. These values, in addition to the total count of gender in the original translations, are presented in Table \ref{tab:prob_differences}. More frequently than in masculine scenarios, the CPD of an original feminine target was negative. This means that the probability for that referent to be translated as feminine instead of masculine was lower but the overall sentence next-token prediction was positive, thus yielding a feminine translation. Figure \ref{fig:violin_prob_difference} further visualises the full distribution of the CPDs.
These results are consistent with prior research showing that models tend to favour masculine translations. In contrast to previous work, this study further quantifies the extent to which \textbf{one translation is more \textit{probable} than the alternative}.

\paragraph{Contrastive Probability Difference per Referent}
We further analyse which \textit{source referent} is translated into which target gender with which probability difference. Table \ref{tab:match_and_mismatch} outlines the number of referents, where their source embedding gender (as outlined in Section \ref{data_filtering}) either matches or mismatches the gender in the target translation, and with what probability difference this has been translated. As an example, the referent `roommate' has a neutral source embedding but has been, for both languages, (majority) translated into feminine with an average CPD of $\sim$0.15 (i.e. 15\% certainty of translating the referent as feminine \textit{instead} of masculine). On average, for both languages, the CPD is slightly higher for matched genders, while slightly more referents mismatch in gender. Matches predominantly occur for masculine-gender inflected referents that have been translated into masculine, or stereotypical feminine referents that have been translated into feminine, while mismatches predominantly occur for referents with a feminine or neutral gender inflection that have been translated into masculine. The model thus shows a higher confidence (higher CPD) when translating masculine or stereotypical referents, and a lower confidence when translating neutral or less stereotypical feminine referents. Appendix \ref{app:prob_diff_referent} presents this in more detail for each referent: Figures \ref{fig:prob_diff_mismatch_referent_ES} and \ref{fig:prob_diff_mismatch_referent_DE} depict mismatches, where the source embedding differs from the target gender. Figures \ref{fig:prob_diff_match_referent_ES} and \ref{fig:prob_diff_match_referent_DE} present matches, where the source embedding matches the target gender.

\begin{table}
\begin{tabular}{c|c|c|c|c|c|c}
 & \textbf{==}& $\overline{\textbf{p. diff}}$ &
  $\sigma$ & \textbf{!=} & $\overline{\textbf{p. diff}}$ & $\sigma$ \\
 \midrule
ES & 80 & .60 & .15 & 90 & .48 & .18 \\
DE & 77 & .53 & .13 & 91 & .43 & .17 \\
\bottomrule
\end{tabular}
\caption{\label{tab:match_and_mismatch}
 Per language, number of referents where the gender of the source embedding matches (==) or mismatches (!=) the gender of the target translation, with the overall average probability difference ($\overline{p. diff}$) and average standard deviation ($\sigma$).
  }
\end{table}


\subsubsection{Linguistic Analysis}
A closer look at salient words reveals a few insights. Looking at source words within the relative top 15\% of saliency scores, 1007 words were salient for EN$\rightarrow$DE, and 1103 for EN$\rightarrow$ES (on average, around one word per sentence). Overall, approximately half of all salient words occur for both language directions. 51\% of all words that were salient for EN$\rightarrow$DE were also salient for EN$\rightarrow$ES, and the other way round, 46\% of all words that were salient for EN$\rightarrow$ES were also salient for EN$\rightarrow$DE. This shows that \textbf{a high number of source words are considered salient for the model, regardless of the target language}.

\begin{figure}[t]
  \includegraphics[width=\columnwidth]{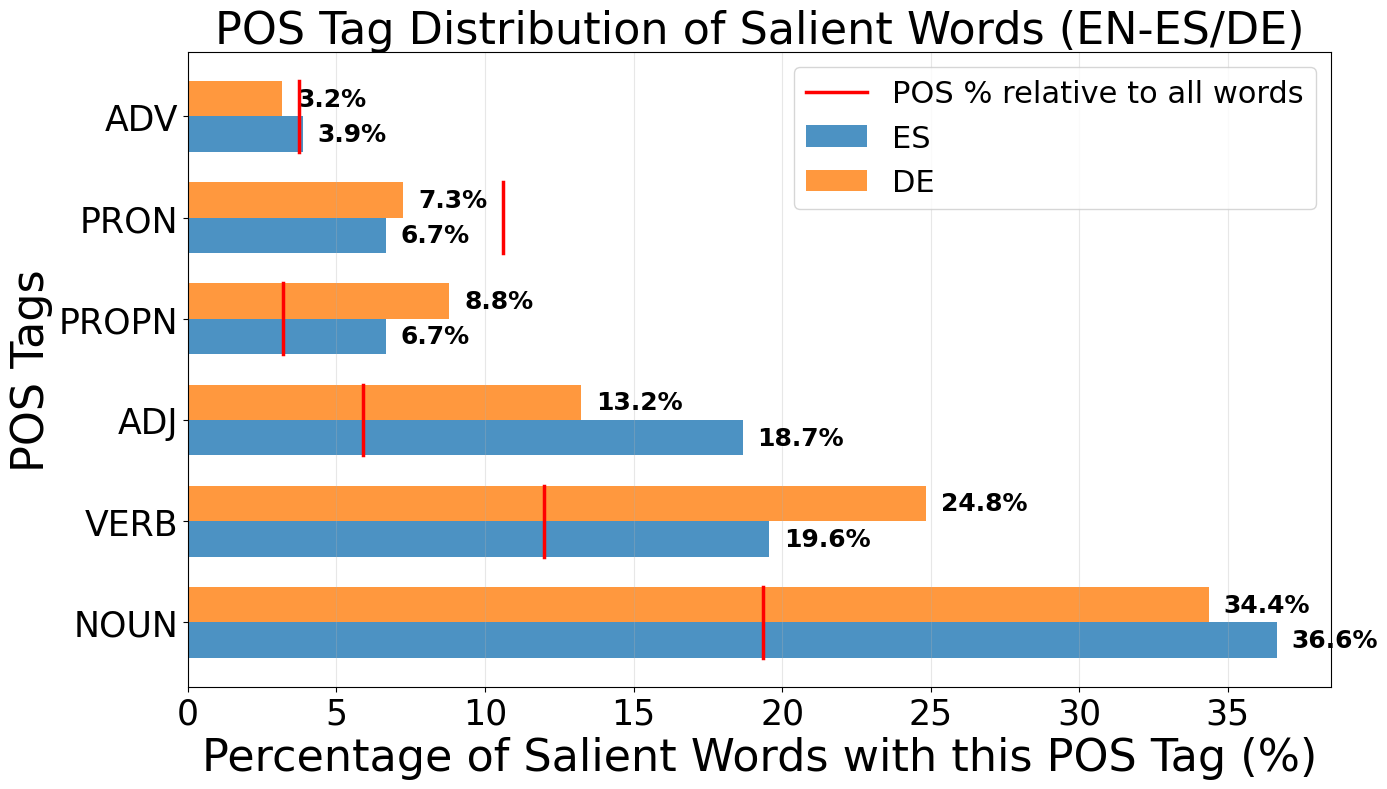}
  \par\vspace{2pt}
  \centering\small (a) Parts-of-speech distribution of salient words.
  \par\vspace{10pt}
  \includegraphics[width=\columnwidth]{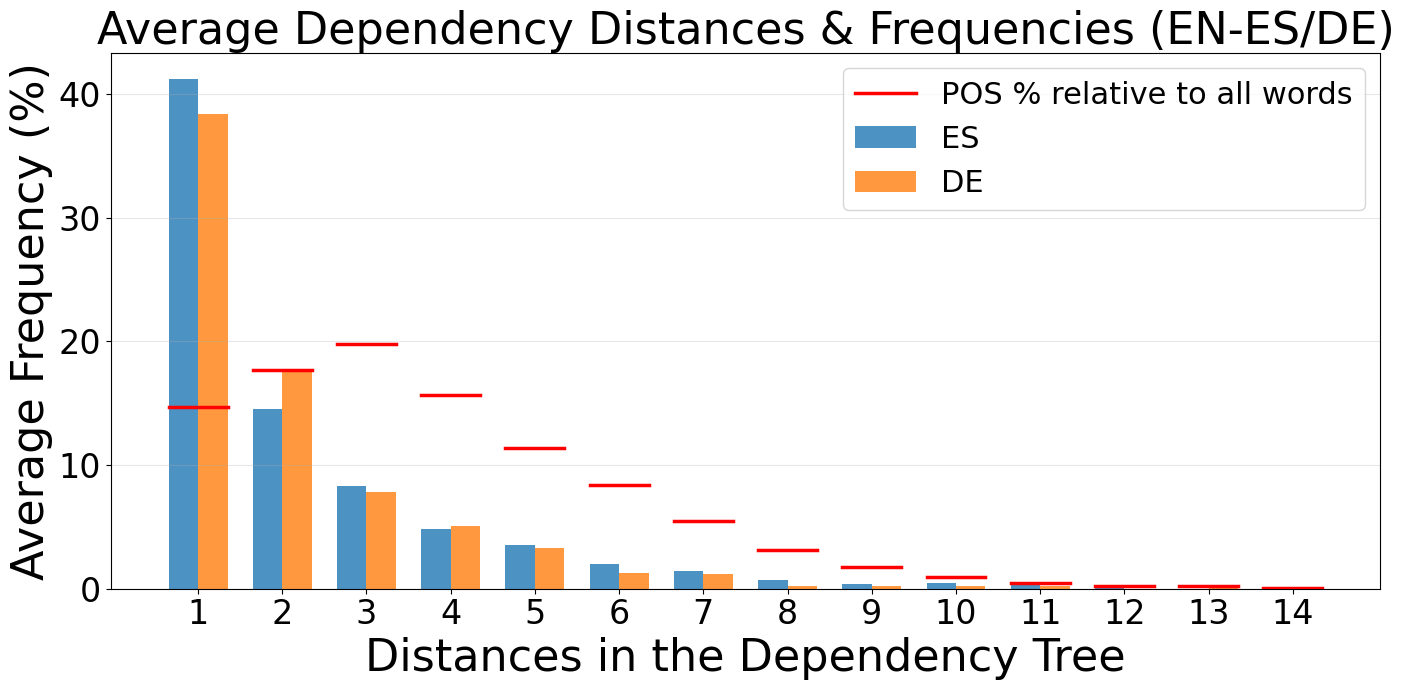}
  \par\vspace{2pt}
  \centering\small (b) Average dependency distances of salient words to referent entity.
  \caption{Parts-of-speech distribution and dependency distances of salient words.}
  \label{fig:linguistic_analysis}
\end{figure}

\paragraph{Parts-of-Speech}
The POS analysis of salient words for both languages is 
similar. This might not be very surprising considering that about half of the salient words overlap for both languages. Both for EN $\rightarrow$ DE/ES, our analysis shows that salient words were of the following POS categories (in descending order): \textbf{nouns} with $\sim$36\% of all salient words (36.6\% for ES; 34.4\% for DE), \textbf{verbs} with $\sim$22\% (ES 19.6\%; DE 24.8\%), \textbf{adjectives} with $\sim$16\% (ES 18.7\%; DE 13.2\%), \textbf{proper nouns} with $\sim$8\% (ES 6.7\%; DE 8.8\%), and \textbf{pronouns} with $\sim$7\% (ES 6.7\%; DE 7.3\%). This is visualised in Figure \ref{fig:linguistic_analysis} (a). The two most predominant categories of nouns and verbs align with findings from Hackenbuchner et al. ~\shortcite{hackenbuchner2025-arxiv}.
The general trend of these salient POS categories follows the relative occurrence of the overall POS frequency of all words, represented by the red vertical lines in Figure \ref{fig:linguistic_analysis} (a). This comparison shows that \textbf{the model considers nouns, adjectives, verbs, and proper nouns to be salient much more frequently than their comparable occurrence in the overall data} (i.e. twice as frequently or even more). Figure \ref{fig:heatmap} in Appendix \ref{app:prob_diff_referent} depicts a heatmap of the POS tags of salient words with respect to each referent for DE and ES.

\begin{table*}[ht!]
\small
\centering
\renewcommand{\arraystretch}{1.4}
\begin{tabularx}{\textwidth}{l l l c c l}
\toprule
\textbf{Sentence} & \textbf{Lang.} & \textbf{Intervention} & \textbf{Gender TR} & \textbf{CPD} & \textbf{Salient word} \\
\midrule

\multirow{6}{*}{\parbox{5.5cm}{\small\textit{``And if they can't get my help, they'll get my \underline{horny} }roommate\textit{ to do it.''}}}
  & \multirow{3}{*}{ES}
    & Original   & \masc{Masc.} & $+0.25$ & \textit{horny} [ADJ, dist: 1] \\
  & & Mask       & \masc{Masc.} & $+0.34$ & $<$mask$>$ \\
  & & Remove     & \masc{Masc.} & $+0.41$ & \\
\cmidrule{2-6}
  & \multirow{3}{*}{DE}
    & Original   & \fem{Fem.}  & $-0.01$ & \textit{horny} [ADJ, dist: 1] \\
  & & Mask       & \fem{Fem.}  & $+0.24$ & $<$mask$>$ \\
  & & Remove     & \masc{Masc.} & $-0.02$ & \\

\midrule

\multirow{6}{*}{\parbox{5.5cm}{\small\textit{``My college }roommate\textit{ and I both had received commitments from our respective \underline{mothers} to fly us out...''}}}
  & \multirow{3}{*}{ES}
    & Original   & \fem{Fem.}  & $+0.20$ & \textit{mothers} [NOUN, dist: 4] \\
  & & Mask       & \masc{Masc.} & $-0.26$ & $<$mask$>$ \\
  & & Flip       & \masc{Masc.} & $-0.13$ & \textit{fathers} [NOUN, dist: 4] \\
\cmidrule{2-6}
  & \multirow{3}{*}{DE}
    & Original   & \fem{Fem.}  & $+0.33$ & \textit{mothers} [NOUN, dist: 4] \\
  & & Mask       & \fem{Fem.}  & $0.15$ & $<$mask$>$ \\
  & & Flip       & \masc{Masc.} & $+0.10$ & \textit{fathers} [NOUN, dist: 4] \\

\bottomrule
\end{tabularx}
\caption{Qualitative example: gender translation analysis with salient word intervention of either removing, masking of flipping salient words. ``CPD'' stands for contrastive probability difference, ``Gender TR'' for gender in translation. Colours indicate the gender of the referent in the target translation: \masc{Masc.} for masculine, \fem{Fem} for feminine.}
\label{tab:implications-analysis}
\end{table*}

\paragraph{Dependency Distance}
As for the dependency distance analysis, both languages reveal very similar patterns in terms of the grammatical structure of salient words in relation to the target referent. The predominant syntactic dependency distance between salient words and a target referent is \textbf{1} (41.2\% for ES; 38.3\% for DE). To a lesser extent, salient words are found at a dependency distance of \textbf{2} away from the referent (14.5\% for ES; 17.7\% for DE) or at a distance of \textbf{3} away (8.3\% for ES; 7.8\% for DE). The following distances decrease in order and become negligible from a distance of 4 onwards. This is visualised in Figure \ref{fig:linguistic_analysis} (b). These results similarly align with findings from Hackenbuchner et al. ~\shortcite{hackenbuchner2025-arxiv}.

This finding shows that in the referent translation of gender, \textbf{the model is predominantly influenced by words that are grammatically and structurally close}, with no more than a dependency distance of 3 away. The distributed dependency distances between the target word and all other words in each sentence is depicted by the horizontal red lines in Figure \ref{fig:linguistic_analysis} (b). This comparison shows that even though the most frequent dependency distance in the overall data is distance 3, the model clearly finds words at a distance of 1 to be much more salient (i.e. almost three times as frequently).

\subsubsection{Implications} 

The contrastive probability differences in this study showed that the model was more confident in opting for a masculine gender in translation instead of a feminine one. 
Salient cues affecting the model's choice of gender in translation were found to predominantly be nouns, verbs, and adjectives, and were predominantly at a distance of 1 or 2 away from the referent. 
Based on these findings, in future work we can explore the impact of specific content on gender inflection in a more targeted manner, for example by testing if removing, masking, or changing salient words actually has an impact on the gender in the target translation and on the CPD in each language.

We illustrate this direction with a qualitative example involving the neutrally associated referent \textit{roommate}, which was translated as feminine on four occasions and as masculine on two occasions in both ES and DE. In Table \ref{tab:implications-analysis}, we look at two examples: in the first, the salient word is at a dependency distance of 1 away from the referent, while in the second, the salient word is a noun. We examine what influence removing, masking or flipping salient words has on the gender in the target translation and on the CPD for each language.
These preliminary examples suggest that \textbf{words considered salient by the model have a direct effect on the gender in translation}. In future work, we aim to build on the research presented in this paper by 
scaling this analysis to measure the causal influence of salient words on the choice of gender in translation.



\section{Conclusion}\label{conclusion}
In this paper, we presented \textsc{GAND}, a first-of-its-kind large-scale dataset for gender-ambiguous natural data, specifically designed to analyse the influence of contextual cues on gender in translation. \textsc{GAND} has been meticulously curated to stem from different natural data sources, covering different topics and linguistic styles, and to be fully gender-ambiguous w.r.t. a singular referent. We leverage \textsc{GAND-CT} to employ a feature attribution analysis by means of contrastive translations. By analysing salient words, our work shows how specific contextual cues influence gender choices in translation and helps uncover patterns of gender bias in MT.

While we tested two target languages and one NMT model for the task at hand, \textsc{GAND} can be used to benchmark how different MT models handle gender ambiguity across a wide range of target languages. Future work should therefore investigate whether our findings generalise to different sequence-to-sequence models as well as LLMs that support feature attribution analysis.

With this contribution, we highlight the time-consuming manual effort that flows into the creation of high-quality data, both for \textsc{GAND} as well as for \textsc{GAND-CT}. Furthermore, we emphasise the importance of gender-ambiguous natural data in the continued effort to understand and mitigate social biases in MT systems. We encourage further research into gender bias in the translations of ambiguous source texts and into underlying model mechanisms that inform this behaviour.


\section*{Ethical Considerations}\label{ethics}
Understanding and mitigating biased behaviours, such as gendered mistranslations, in MT systems and general-purpose foundation models, requires both mitigation strategies ~\cite{vanmassenhove-etal-2018,saunders-etal-2022,savoldi-etal-2025-decade}, \textit{i.a.}, and interpretability research to understand and reveal mechanisms underlying biased behaviours ~\cite{attanasio-etal-2023-tale,manna-etal-2025-paying,conti-etal-2025-unheard}, \textit{i.a.}.

The gender-ambiguous dataset presented (and the specific referents focused on) in this paper inherently includes all genders. Higher contrastive probability differences towards masculine translations than feminine translations (Section \ref{interpretability}) can lead to representational harm \cite{blodgett-etal-2020}. Source sentences that have been translated into a `neutral' gender in either of the target languages have not been contrasted for analysis (as there
is no contrast and (mis)gendered ‘error detection’ necessary here). Neutral translations for ambiguous scenarios can be considered bias-free ~\cite{Piergentili:23-neutral,savoldi-etal-2025-mgente}.

Given the nature of the online sources from which this dataset was collected, some content may be explicit or disturbing in nature. We strongly advise those reusing or building upon this data to consider the potential impact on researchers and annotators involved in its analysis.

\section*{Limitations}

\paragraph{GAND} In the compilation of \textsc{GAND}, we focus on two natural data sources, C4 and OpenSubtitles, with the aim to broaden the search for topic and linguistic diversity. Nevertheless, we cover only these two sources in the dataset compilation and focus only on English as a source.

\paragraph{Interpretability Analysis} In the subsequent interpretability analysis, we translate and contrast only a subset of 1000 sentences of \textsc{GAND}, which makes up 20\% of all sentences. This limitation is due to the labour-intensive constraints that the gender annotation and contrastive translations methodology adopted in this work pose. For an accurate analysis, this requires gold-standard annotations and contrastive gender alternatives for each sentence. Nevertheless, in future work, we aim to automate this process to scale this methodology and subsequent analysis.

Furthermore, we only conduct translations and analyses on two gender grammatical target languages (Spanish and German) by means of one translation model (OPUS-MT). While the OPUS-MT models are not state-of-the-art in machine translation performance, we selected these for a simple methodological reason: it is one of the few open sequence-to-sequence models available for feature attribution analysis. We do not extend our analysis to general-purpose foundation models capable of translation, which are established in the translation sector, as they represent different mechanisms for text translation, influencing the subsequent attribution analysis. This can be researched in future work.

\section{Acknowledgements}
We would like to thank the anonymous reviewers for their invaluable feedback. This study is part of a strategic basic PhD research (1SH5V24N) fully funded by The Research Foundation – Flanders (FWO) for the time span of four years, from 01.11.2023 until 31.10.2027, and hosted within the Language and Translation Technology Team (LT\textsuperscript{3}) at Ghent University. The computational resources (Stevin Supercomputer Infrastructure) and services used in this work were provided by the VSC (Flemish Supercomputer Center), funded by Ghent University, FWO and the Flemish Government - department EWI.

\bibliography{eamt26}

@inproceedings{attanasio-etal-2023-tale,
    title = "A Tale of Pronouns: Interpretability Informs Gender Bias Mitigation for Fairer Instruction-Tuned Machine Translation",
    author = "Attanasio, Giuseppe  and
      Plaza del Arco, Flor Miriam  and
      Nozza, Debora  and
      Lauscher, Anne",
    editor = "Bouamor, Houda  and
      Pino, Juan  and
      Bali, Kalika",
    booktitle = "Proceedings of the 2023 Conference on Empirical Methods in Natural Language Processing",
    month = dec,
    year = "2023",
    address = "Singapore",
    publisher = "Association for Computational Linguistics",
    url = "https://aclanthology.org/2023.emnlp-main.243/",
    doi = "10.18653/v1/2023.emnlp-main.243",
    pages = "3996--4014"
}

@inproceedings{piergentili-etal-2023-hi,
    title = "Hi Guys or Hi Folks? Benchmarking Gender-Neutral Machine Translation with the {G}e{NTE} Corpus",
    author = "Piergentili, Andrea  and
      Savoldi, Beatrice  and
      Fucci, Dennis  and
      Negri, Matteo  and
      Bentivogli, Luisa",
    editor = "Bouamor, Houda  and
      Pino, Juan  and
      Bali, Kalika",
    booktitle = "Proceedings of the 2023 Conference on Empirical Methods in Natural Language Processing",
    month = dec,
    year = "2023",
    address = "Singapore",
    publisher = "Association for Computational Linguistics",
    url = "https://aclanthology.org/2023.emnlp-main.873/",
    doi = "10.18653/v1/2023.emnlp-main.873",
    pages = "14124--14140",
}

@inproceedings{yin-neubig-2022-interpreting,
    title = "Interpreting Language Models with Contrastive Explanations",
    author = "Yin, Kayo  and
      Neubig, Graham",
    editor = "Goldberg, Yoav  and
      Kozareva, Zornitsa  and
      Zhang, Yue",
    booktitle = "Proceedings of the 2022 Conference on Empirical Methods in Natural Language Processing",
    month = dec,
    year = "2022",
    address = "Abu Dhabi, United Arab Emirates",
    publisher = "Association for Computational Linguistics",
    url = "https://aclanthology.org/2022.emnlp-main.14/",
    doi = "10.18653/v1/2022.emnlp-main.14",
    pages = "184--198"
}

@inproceedings{manna-etal-2025-paying,
    title = "Are We Paying Attention to Her? Investigating Gender Disambiguation and Attention in Machine Translation",
    author = "Manna, Chiara  and
      Alishahi, Afra  and
      Blain, Fr{\'e}d{\'e}ric  and
      Vanmassenhove, Eva",
    editor = "Hackenbuchner, Jani{\c{c}}a  and
      Bentivogli, Luisa  and
      Daems, Joke  and
      Manna, Chiara  and
      Savoldi, Beatrice  and
      Vanmassenhove, Eva",
    booktitle = "Proceedings of the 3rd Workshop on Gender-Inclusive Translation Technologies (GITT 2025)",
    month = jun,
    year = "2025",
    address = "Geneva, Switzerland",
    publisher = "European Association for Machine Translation",
    url = "https://aclanthology.org/2025.gitt-1.1/",
    pages = "1--16",
    ISBN = "978-2-9701897-4-9"
}

@inproceedings{sarti-etal-2023-inseq,
    title = "Inseq: An Interpretability Toolkit for Sequence Generation Models",
    author = "Sarti, Gabriele  and
      Feldhus, Nils  and
      Sickert, Ludwig  and
      van der Wal, Oskar",
    editor = "Bollegala, Danushka  and
      Huang, Ruihong  and
      Ritter, Alan",
    booktitle = "Proceedings of the 61st Annual Meeting of the Association for Computational Linguistics (Volume 3: System Demonstrations)",
    month = jul,
    year = "2023",
    address = "Toronto, Canada",
    publisher = "Association for Computational Linguistics",
    url = "https://aclanthology.org/2023.acl-demo.40/",
    doi = "10.18653/v1/2023.acl-demo.40",
    pages = "421--435"
}

@article{savoldi-etal-2025-decade,
    author = "Savoldi, Beatrice and Bastings, Jasmijn and Bentivogli, Luisa and Vanmassenhove, Eva",
    title = "A decade of gender bias in machine translation",
    journal = "Patterns",
    volume = "6",
    issue = "6",
    doi = "10.1016/j.patter.2025.101257",
    url = "https://doi.org/10.1016/j.patter.2025.101257", 
    year = "2025"
}

@misc{ferrando2024,
      title={A Primer on the Inner Workings of Transformer-based Language Models}, 
      author={Javier Ferrando and Gabriele Sarti and Arianna Bisazza and Marta R. Costa-jussà},
      year={2024},
      eprint={2405.00208},
      archivePrefix={arXiv},
      primaryClass={cs.CL},
      url={https://arxiv.org/abs/2405.00208}, 
}

@inproceedings{vanmassenhove-etal-2018,
    title = "Getting Gender Right in Neural Machine Translation",
    author = "Vanmassenhove, Eva  and
      Hardmeier, Christian  and
      Way, Andy",
    editor = "Riloff, Ellen  and
      Chiang, David  and
      Hockenmaier, Julia  and
      Tsujii, Jun{'}ichi",
    booktitle = "Proceedings of the 2018 Conference on Empirical Methods in Natural Language Processing",
    month = oct # "-" # nov,
    year = "2018",
    address = "Brussels, Belgium",
    publisher = "Association for Computational Linguistics",
    url = "https://aclanthology.org/D18-1334/",
    doi = "10.18653/v1/D18-1334",
    pages = "3003--3008"
}

@inproceedings{saunders-etal-2022,
    title = "First the Worst: Finding Better Gender Translations During Beam Search",
    author = "Saunders, Danielle  and
      Sallis, Rosie  and
      Byrne, Bill",
    editor = "Muresan, Smaranda  and
      Nakov, Preslav  and
      Villavicencio, Aline",
    booktitle = "Findings of the Association for Computational Linguistics: ACL 2022",
    month = may,
    year = "2022",
    address = "Dublin, Ireland",
    publisher = "Association for Computational Linguistics",
    url = "https://aclanthology.org/2022.findings-acl.301/",
    doi = "10.18653/v1/2022.findings-acl.301",
    pages = "3814--3823"
}

@misc{räuker2023,
      title={Toward Transparent AI: A Survey on Interpreting the Inner Structures of Deep Neural Networks}, 
      author={Tilman Räuker and Anson Ho and Stephen Casper and Dylan Hadfield-Menell},
      year={2023},
      eprint={2207.13243},
      archivePrefix={arXiv},
      primaryClass={cs.LG},
      url={https://arxiv.org/abs/2207.13243}, 
}

@inproceedings{conti-etal-2025-unheard,
    title = "The Unheard Alternative: Contrastive Explanations for Speech-to-Text Models",
    author = "Conti, Lina  and
      Fucci, Dennis  and
      Gaido, Marco  and
      Negri, Matteo  and
      Wisniewski, Guillaume  and
      Bentivogli, Luisa",
    editor = "Belinkov, Yonatan  and
      Mueller, Aaron  and
      Kim, Najoung  and
      Mohebbi, Hosein  and
      Chen, Hanjie  and
      Arad, Dana  and
      Sarti, Gabriele",
    booktitle = "Proceedings of the 8th BlackboxNLP Workshop: Analyzing and Interpreting Neural Networks for NLP",
    month = nov,
    year = "2025",
    address = "Suzhou, China",
    publisher = "Association for Computational Linguistics",
    url = "https://aclanthology.org/2025.blackboxnlp-1.23/",
    doi = "10.18653/v1/2025.blackboxnlp-1.23",
    pages = "398--414",
    ISBN = "979-8-89176-346-3"
}

@inproceedings{tiedemann2020,
  author = {J{\"o}rg Tiedemann and Santhosh Thottingal},
  title = {{OPUS-MT} — {B}uilding open translation services for the {W}orld},
  booktitle = {Proceedings of the 22nd Annual Conferenec of the European Association for Machine Translation (EAMT)},
  url="https://aclanthology.org/2020.eamt-1.61/",
  year = {2020},
  address = {Lisbon, Portugal}
 }

@article{hackenbuchner-etal-2025-clin,
    author = "Hackenbuchner, Jani{\c{c}}a and Tezcan, Arda and Daems, Joke",
    title = "Gender Bias and the Role of Context in Human Perception and Machine Translation",
    journal = "Computational Linguistics in the Netherlands Journal",
    volume = "14",
    url = "https://www.clinjournal.org/clinj/article/view/197", 
    year = "2025"
}

@inproceedings{hackenbuchner-etal-2024-project,
    title = "Automatic detection of (potential) factors in the source text leading to gender bias in machine translation",
    author = "Hackenbuchner, Jani{\c{c}}a  and
      Tezcan, Arda  and
      Daems, Joke",
    editor = "Scarton, Carolina  and
      Prescott, Charlotte  and
      Bayliss, Chris  and
      Oakley, Chris  and
      Wright, Joanna  and
      Wrigley, Stuart  and
      Song, Xingyi  and
      Gow-Smith, Edward  and
      Forcada, Mikel  and
      Moniz, Helena",
    booktitle = "Proceedings of the 25th Annual Conference of the European Association for Machine Translation (Volume 2)",
    month = jun,
    year = "2024",
    address = "Sheffield, UK",
    publisher = "European Association for Machine Translation (EAMT)",
    url = "https://aclanthology.org/2024.eamt-2.14/",
    pages = "27--28"
}

@inproceedings{hackenbuchner-etal-2025-genderous,
	title = {{GENDEROUS}: {Machine} {Translation} and {Cross}-{Linguistic} {Evaluation} of a {Gender}-{Ambiguous} {Dataset}},
	language = {en},
    year={2025},
    booktitle={Proceedings of the 6th Workshop on Gender Bias in Natural Language Processing (GeBNLP)},
    pages={302–-319},
    url="https://aclanthology.org/2025.gebnlp-1.27/",
    publisher={Association for Computational Linguistics},
	author = {Hackenbuchner, Janiça and Gkovedarou, Eleni and Daems, Joke}
}

@inproceedings{caliskan-etal-2022,
	address = {Oxford United Kingdom},
	title = {Gender {Bias} in {Word} {Embeddings}: {A} {Comprehensive} {Analysis} of {Frequency}, {Syntax}, and {Semantics}},
	isbn = {978-1-4503-9247-1},
	shorttitle = {Gender {Bias} in {Word} {Embeddings}},
	url = {https://dl.acm.org/doi/10.1145/3514094.3534162},
	doi = {10.1145/3514094.3534162},
	language = {en},
	urldate = {2025-09-25},
	booktitle = {Proceedings of the 2022 {AAAI}/{ACM} {Conference} on {AI}, {Ethics}, and {Society}},
	publisher = {ACM},
	author = {Caliskan, Aylin and Ajay, Pimparkar Parth and Charlesworth, Tessa and Wolfe, Robert and Banaji, Mahzarin R.},
	month = jul,
	year = {2022},
	pages = {156--170}
}

@article{Bolukbasi-etal-2016,
	title = {Man is to {Computer} {Programmer} as {Woman} is to {Homemaker}? {Debiasing} {Word} {Embeddings}},
	volume = {abs/1607.06520},
	journal = {CoRR},
    url="https://dl.acm.org/doi/10.5555/3157382.3157584",
	author = {Bolukbasi, Tolga and Chang, Kai-Wei and Zou, James and Saligrama, Venkatesh and Kalai, Adam},
	year = {2016},
}

@inproceedings{kocmi-etal-2020,
  title={Gender Coreference and Bias Evaluation at WMT 2020},
  author={Kocmi, Tom and Limisiewicz, Tomasz and Stanovsky, Gabriel},
  booktitle={Proceedings of the 5th Conference on Machine Translation (WMT)},
  publisher={Association for Computational Linguistics},
  pages={357–364},
  year={2020},
  url="https://aclanthology.org/2020.wmt-1.39/"
}

@inproceedings{levy-etal-2021,
	address = {Punta Cana, Dominican Republic},
	title = {Collecting a {Large}-{Scale} {Gender} {Bias} {Dataset} for {Coreference} {Resolution} and {Machine} {Translation}},
	url = {https://aclanthology.org/2021.findings-emnlp.211},
	doi = {10.18653/v1/2021.findings-emnlp.211},
	language = {en},
	urldate = {2025-09-25},
	booktitle = {Findings of the {Association} for {Computational} {Linguistics}: {EMNLP} 2021},
	publisher = {Association for Computational Linguistics},
	author = {Levy, Shahar and Lazar, Koren and Stanovsky, Gabriel},
	year = {2021},
	pages = {2470--2480}
}

@incollection{vanmassenhove-2024,
  title={Gender bias in machine translation and the era of large language models},
  author={Vanmassenhove, Eva},
  booktitle={Gendered Technology in Translation and Interpreting},
  pages={225--252},
  year={2024},
  publisher={Routledge},
  url="https://www.taylorfrancis.com/chapters/edit/10.4324/9781003465508-12/gender-bias-machine-translation-era-large-language-models-eva-vanmassenhove"
}

@inproceedings{savoldi-etal-2025-mgente,
    title = "Mind the Inclusivity Gap: Multilingual Gender-Neutral Translation Evaluation with m{G}e{NTE}",
    author = "Savoldi, Beatrice  and
      Attanasio, Giuseppe  and
      Cupin, Eleonora  and
      Gkovedarou, Eleni  and
      Hackenbuchner, Jani{\c{c}}a  and
      Lauscher, Anne  and
      Negri, Matteo  and
      Piergentili, Andrea  and
      Thind, Manjinder  and
      Bentivogli, Luisa",
    editor = "Christodoulopoulos, Christos  and
      Chakraborty, Tanmoy  and
      Rose, Carolyn  and
      Peng, Violet",
    booktitle = "Proceedings of the 2025 Conference on Empirical Methods in Natural Language Processing",
    month = nov,
    year = "2025",
    address = "Suzhou, China",
    publisher = "Association for Computational Linguistics",
    url = "https://aclanthology.org/2025.emnlp-main.692/",
    doi = "10.18653/v1/2025.emnlp-main.692",
    pages = "13709--13731",
    ISBN = "979-8-89176-332-6"
}

@misc{hackenbuchner2025-arxiv,
      title={What Triggers my Model? Contrastive Explanations Inform Gender Choices by Translation Models}, 
      author={Janiça Hackenbuchner and Arda Tezcan and Joke Daems},
      year={2025},
      eprint={2512.08440},
      archivePrefix={arXiv},
      primaryClass={cs.CL},
      url={https://arxiv.org/abs/2512.08440}, 
}

@inproceedings{renduchintala-etal-2021,
    title = "Gender bias amplification during Speed-Quality optimization in Neural Machine Translation",
    author = "Renduchintala, Adithya  and
      Diaz, Denise  and
      Heafield, Kenneth  and
      Li, Xian  and
      Diab, Mona",
    editor = "Zong, Chengqing  and
      Xia, Fei  and
      Li, Wenjie  and
      Navigli, Roberto",
    booktitle = "Proceedings of the 59th Annual Meeting of the Association for Computational Linguistics and the 11th International Joint Conference on Natural Language Processing (Volume 2: Short Papers)",
    month = aug,
    year = "2021",
    address = "Online",
    publisher = "Association for Computational Linguistics",
    url = "https://aclanthology.org/2021.acl-short.15/",
    doi = "10.18653/v1/2021.acl-short.15",
    pages = "99--109"
}

@inproceedings{robinson-etal-2024,
    title = "{M}i{TT}en{S}: A Dataset for Evaluating Gender Mistranslation",
    author = "Robinson, Kevin  and
      Kudugunta, Sneha  and
      Stella, Romina  and
      Dev, Sunipa  and
      Bastings, Jasmijn",
    editor = "Al-Onaizan, Yaser  and
      Bansal, Mohit  and
      Chen, Yun-Nung",
    booktitle = "Proceedings of the 2024 Conference on Empirical Methods in Natural Language Processing",
    month = nov,
    year = "2024",
    address = "Miami, Florida, USA",
    publisher = "Association for Computational Linguistics",
    url = "https://aclanthology.org/2024.emnlp-main.238/",
    doi = "10.18653/v1/2024.emnlp-main.238",
    pages = "4115--4124"
}

@inproceedings{zhao-etal-2018-gender,
    title = "Gender Bias in Coreference Resolution: Evaluation and Debiasing Methods",
    author = "Zhao, Jieyu  and
      Wang, Tianlu  and
      Yatskar, Mark  and
      Ordonez, Vicente  and
      Chang, Kai-Wei",
    editor = "Walker, Marilyn  and
      Ji, Heng  and
      Stent, Amanda",
    booktitle = "Proceedings of the 2018 Conference of the North {A}merican Chapter of the Association for Computational Linguistics: Human Language Technologies, Volume 2 (Short Papers)",
    month = jun,
    year = "2018",
    address = "New Orleans, Louisiana",
    publisher = "Association for Computational Linguistics",
    url = "https://aclanthology.org/N18-2003/",
    doi = "10.18653/v1/N18-2003",
    pages = "15--20"
}

@inproceedings{habash-etal-2019,
    title = "Automatic Gender Identification and Reinflection in {A}rabic",
    author = "Habash, Nizar  and
      Bouamor, Houda  and
      Chung, Christine",
    editor = "Costa-juss{\`a}, Marta R.  and
      Hardmeier, Christian  and
      Radford, Will  and
      Webster, Kellie",
    booktitle = "Proceedings of the First Workshop on Gender Bias in Natural Language Processing",
    month = aug,
    year = "2019",
    address = "Florence, Italy",
    publisher = "Association for Computational Linguistics",
    url = "https://aclanthology.org/W19-3822/",
    doi = "10.18653/v1/W19-3822",
    pages = "155--165"
}

@inproceedings{rarrick-etal-2023,
author = {Rarrick, Spencer and Naik, Ranjita and Mathur, Varun and Poudel, Sundar and Chowdhary, Vishal},
title = {GATE: A Challenge Set for Gender-Ambiguous Translation Examples},
year = {2023},
isbn = {9798400702310},
publisher = {Association for Computing Machinery},
address = {New York, NY, USA},
url = {https://doi.org/10.1145/3600211.3604675},
doi = {10.1145/3600211.3604675},
booktitle = {Proceedings of the 2023 AAAI/ACM Conference on AI, Ethics, and Society},
pages = {845–854},
numpages = {10},
location = {Montr\'{e}al, QC, Canada},
series = {AIES '23}
}

@inproceedings{wisniewski-etal-2022,
    title = "Analyzing Gender Translation Errors to Identify Information Flows between the Encoder and Decoder of a {NMT} System",
    author = "Wisniewski, Guillaume  and
      Zhu, Lichao  and
      Ballier, Nicolas  and
      Yvon, Fran{\c{c}}ois",
    editor = "Bastings, Jasmijn  and
      Belinkov, Yonatan  and
      Elazar, Yanai  and
      Hupkes, Dieuwke  and
      Saphra, Naomi  and
      Wiegreffe, Sarah",
    booktitle = "Proceedings of the Fifth BlackboxNLP Workshop on Analyzing and Interpreting Neural Networks for NLP",
    month = dec,
    year = "2022",
    address = "Abu Dhabi, United Arab Emirates (Hybrid)",
    publisher = "Association for Computational Linguistics",
    url = "https://aclanthology.org/2022.blackboxnlp-1.13/",
    doi = "10.18653/v1/2022.blackboxnlp-1.13",
    pages = "153--163"
}

@inproceedings{mastromichalakis-etal-2025,
    title = "Assumed Identities: Quantifying Gender Bias in Machine Translation of Gender-Ambiguous Occupational Terms",
    author = "Mastromichalakis, Orfeas Menis and
      Filandrianos, Giorgos  and
      Symeonaki, Maria  and
      Stamou, Giorgos",
    editor = "Christodoulopoulos, Christos  and
      Chakraborty, Tanmoy  and
      Rose, Carolyn  and
      Peng, Violet",
    booktitle = "Proceedings of the 2025 Conference on Empirical Methods in Natural Language Processing",
    month = nov,
    year = "2025",
    address = "Suzhou, China",
    publisher = "Association for Computational Linguistics",
    url = "https://aclanthology.org/2025.emnlp-main.1640/",
    doi = "10.18653/v1/2025.emnlp-main.1640",
    pages = "32221--32237",
    ISBN = "979-8-89176-332-6"
}

@misc{conti-etal-2025,
      title={Voice, Bias, and Coreference: An Interpretability Study of Gender in Speech Translation}, 
      author={Lina Conti and Dennis Fucci and Marco Gaido and Matteo Negri and Guillaume Wisniewski and Luisa Bentivogli},
      year={2025},
      eprint={2511.21517},
      archivePrefix={arXiv},
      primaryClass={cs.CL},
      url={https://arxiv.org/abs/2511.21517}, 
}

@article{savoldi-etal-2021,
    title = "Gender Bias in Machine Translation",
    author = "Savoldi, Beatrice  and
      Gaido, Marco  and
      Bentivogli, Luisa  and
      Negri, Matteo  and
      Turchi, Marco",
    editor = "Roark, Brian  and
      Nenkova, Ani",
    journal = "Transactions of the Association for Computational Linguistics",
    volume = "9",
    year = "2021",
    address = "Cambridge, MA",
    publisher = "MIT Press",
    url = "https://aclanthology.org/2021.tacl-1.51/",
    doi = "10.1162/tacl_a_00401",
    pages = "845--874"
}

@inproceedings{blodgett-etal-2020,
	address = {Online},
	title = {Language ({Technology}) is {Power}: {A} {Critical} {Survey} of “{Bias}” in {NLP}},
	shorttitle = {Language ({Technology}) is {Power}},
	url = {https://aclanthology.org/2020.acl-main.485/},
	doi = {10.18653/v1/2020.acl-main.485},
	urldate = {2025-03-27},
	booktitle = {Proceedings of the 58th {Annual} {Meeting} of the {Association} for {Computational} {Linguistics}},
	publisher = {Association for Computational Linguistics},
	author = {Blodgett, Su Lin and Barocas, Solon and Daumé III, Hal and Wallach, Hanna},
	editor = {Jurafsky, Dan and Chai, Joyce and Schluter, Natalie and Tetreault, Joel},
	month = jul,
	year = {2020},
	pages = {5454--5476},
}

@article{costa-jussa-etal-2022,
	title = {Interpreting {Gender} {Bias} in {Neural} {Machine} {Translation}: {Multilingual} {Architecture} {Matters}},
	volume = {36},
	issn = {2374-3468, 2159-5399},
	shorttitle = {Interpreting {Gender} {Bias} in {Neural} {Machine} {Translation}},
	url = {https://ojs.aaai.org/index.php/AAAI/article/view/21442},
	doi = {10.1609/aaai.v36i11.21442},
	language = {en},
	number = {11},
	urldate = {2025-09-30},
	journal = {Proceedings of the AAAI Conference on Artificial Intelligence},
	author = {Costa-jussà, Marta R. and Escolano, Carlos and Basta, Christine and Ferrando, Javier and Batlle, Roser and Kharitonova, Ksenia},
	month = jun,
	year = {2022},
	pages = {11855--11863}
}

@misc{lage-etal-2019,
	title = {An {Evaluation} of the {Human}-{Interpretability} of {Explanation}},
	url = {http://arxiv.org/abs/1902.00006},
	doi = {10.48550/arXiv.1902.00006},
	language = {en},
	urldate = {2025-09-30},
	publisher = {arXiv},
	author = {Lage, Isaac and Chen, Emily and He, Jeffrey and Narayanan, Menaka and Kim, Been and Gershman, Sam and Doshi-Velez, Finale},
	month = aug,
	year = {2019},
	note = {arXiv:1902.00006 [cs]}
}

@inproceedings{vamvas-sennrich-2021,
	address = {Online and Punta Cana, Dominican Republic},
	title = {Contrastive {Conditioning} for {Assessing} {Disambiguation} in {MT}: {A} {Case} {Study} of {Distilled} {Bias}},
	shorttitle = {Contrastive {Conditioning} for {Assessing} {Disambiguation} in {MT}},
	url = {https://aclanthology.org/2021.emnlp-main.803},
	doi = {10.18653/v1/2021.emnlp-main.803},
	language = {en},
	urldate = {2025-09-30},
	booktitle = {Proceedings of the 2021 {Conference} on {Empirical} {Methods} in {Natural} {Language} {Processing}},
	publisher = {Association for Computational Linguistics},
	author = {Vamvas, Jannis and Sennrich, Rico},
	year = {2021},
	pages = {10246--10265}
}

@inproceedings{saunders-olsen-2023,
    title = "Gender, names and other mysteries: Towards the ambiguous for gender-inclusive translation",
    author = "Saunders, Danielle  and
      Olsen, Katrina",
    editor = "Vanmassenhove, Eva  and
      Savoldi, Beatrice  and
      Bentivogli, Luisa  and
      Daems, Joke  and
      Hackenbuchner, Jani{\c{c}}a",
    booktitle = "Proceedings of the First Workshop on Gender-Inclusive Translation Technologies",
    month = jun,
    year = "2023",
    address = "Tampere, Finland",
    publisher = "European Association for Machine Translation",
    url = "https://aclanthology.org/2023.gitt-1.8/",
    pages = "85--93"
}

@inproceedings{bentivogli-etal-2020,
    title = "Gender in Danger? Evaluating Speech Translation Technology on the {M}u{ST}-{SHE} Corpus",
    author = "Bentivogli, Luisa  and
      Savoldi, Beatrice  and
      Negri, Matteo  and
      Di Gangi, Mattia A.  and
      Cattoni, Roldano  and
      Turchi, Marco",
    editor = "Jurafsky, Dan  and
      Chai, Joyce  and
      Schluter, Natalie  and
      Tetreault, Joel",
    booktitle = "Proceedings of the 58th Annual Meeting of the Association for Computational Linguistics",
    month = jul,
    year = "2020",
    address = "Online",
    publisher = "Association for Computational Linguistics",
    url = "https://aclanthology.org/2020.acl-main.619/",
    doi = "10.18653/v1/2020.acl-main.619",
    pages = "6923--6933"
}

@inproceedings{saunders-byrne-2020,
	address = {Online},
	title = {Reducing {Gender} {Bias} in {Neural} {Machine} {Translation} as a {Domain} {Adaptation} {Problem}},
	url = {https://www.aclweb.org/anthology/2020.acl-main.690},
	doi = {10.18653/v1/2020.acl-main.690},
	language = {en},
	urldate = {2025-09-25},
	booktitle = {Proceedings of the 58th {Annual} {Meeting} of the {Association} for {Computational} {Linguistics}},
	publisher = {Association for Computational Linguistics},
	author = {Saunders, Danielle and Byrne, Bill},
	year = {2020},
	pages = {7724--7736},
}

@inproceedings{filandrianos-etal-2025,
    title = "{GAMBIT}+: A Challenge Set for Evaluating Gender Bias in Machine Translation Quality Estimation Metrics",
    author = "Filandrianos, George  and
      Menis Mastromichalakis, Orfeas  and
      Mohammed, Wafaa  and
      Attanasio, Giuseppe  and
      Zerva, Chrysoula",
    editor = "Haddow, Barry  and
      Kocmi, Tom  and
      Koehn, Philipp  and
      Monz, Christof",
    booktitle = "Proceedings of the Tenth Conference on Machine Translation",
    month = nov,
    year = "2025",
    address = "Suzhou, China",
    publisher = "Association for Computational Linguistics",
    url = "https://aclanthology.org/2025.wmt-1.19/",
    doi = "10.18653/v1/2025.wmt-1.19",
    pages = "314--326",
    ISBN = "979-8-89176-341-8"
}

@inproceedings{stanovsky-etal-2019,
	address = {Florence, Italy},
	title = {Evaluating {Gender} {Bias} in {Machine} {Translation}},
	url = {https://www.aclweb.org/anthology/P19-1164},
	doi = {10.18653/v1/P19-1164},
	language = {en},
	urldate = {2025-09-25},
	booktitle = {Proceedings of the 57th {Annual} {Meeting} of the {Association} for {Computational} {Linguistics}},
	publisher = {Association for Computational Linguistics},
	author = {Stanovsky, Gabriel and Smith, Noah A. and Zettlemoyer, Luke},
	year = {2019},
	pages = {1679--1684}
}

@inproceedings{kotek-etal-2023,
	address = {Delft Netherlands},
	title = {Gender bias and stereotypes in {Large} {Language} {Models}},
	isbn = {979-8-4007-0113-9},
	url = {https://dl.acm.org/doi/10.1145/3582269.3615599},
	doi = {10.1145/3582269.3615599},
	language = {en},
	urldate = {2025-09-25},
	booktitle = {Proceedings of {The} {ACM} {Collective} {Intelligence} {Conference}},
	publisher = {ACM},
	author = {Kotek, Hadas and Dockum, Rikker and Sun, David},
	month = nov,
	year = {2023},
	pages = {12--24}
}

@inproceedings{currey-etal-2022,
    title = "{MT}-{G}en{E}val: A Counterfactual and Contextual Dataset for Evaluating Gender Accuracy in Machine Translation",
    author = "Currey, Anna  and
      Nadejde, Maria  and
      Pappagari, Raghavendra Reddy  and
      Mayer, Mia  and
      Lauly, Stanislas  and
      Niu, Xing  and
      Hsu, Benjamin  and
      Dinu, Georgiana",
    editor = "Goldberg, Yoav  and
      Kozareva, Zornitsa  and
      Zhang, Yue",
    booktitle = "Proceedings of the 2022 Conference on Empirical Methods in Natural Language Processing",
    month = dec,
    year = "2022",
    address = "Abu Dhabi, United Arab Emirates",
    publisher = "Association for Computational Linguistics",
    url = "https://aclanthology.org/2022.emnlp-main.288/",
    doi = "10.18653/v1/2022.emnlp-main.288",
    pages = "4287--4299",
}

@inproceedings{gkovedarou_gender_2025,
	title = {Gender {Bias} in {English}-to-{Greek} {Machine} {Translation}},
	url = {https://aclanthology.org/2025.gitt-1.2/},
	language = {en},
    booktitle={Proceedings of the 3rd Workshop on Gender-Inclusive Translation Technologies (GITT 2025)},
	publisher = {European Association for Machine Translation},
	author = {Gkovedarou, Eleni and Daems, Joke and Bruyne, Luna De},
	year = {2025},
    pages={17-–45},
    address={Geneva, Switzerland}
}

@inproceedings{Piergentili:23-neutral,
    title = "Gender Neutralization for an Inclusive Machine Translation: from Theoretical Foundations to Open Challenges",
    author = "Piergentili, Andrea  and
      Fucci, Dennis  and
      Savoldi, Beatrice  and
      Bentivogli, Luisa  and
      Negri, Matteo",
    editor = "Vanmassenhove, Eva  and
      Savoldi, Beatrice  and
      Bentivogli, Luisa  and
      Daems, Joke  and
      Hackenbuchner, Jani{\c{c}}a",
    booktitle = "Proceedings of the First Workshop on Gender-Inclusive Translation Technologies",
    month = jun,
    year = "2023",
    address = "Tampere, Finland",
    publisher = "European Association for Machine Translation",
    url = "https://aclanthology.org/2023.gitt-1.7/",
    pages = "71--83",
}

@inproceedings{pranav-etal-2025,
    title = "Glitter: A Multi-Sentence, Multi-Reference Benchmark for Gender-Fair {G}erman Machine Translation",
    author = "Pranav, A  and
      Hackenbuchner, Jani{\c{c}}a  and
      Attanasio, Giuseppe  and
      Lardelli, Manuel  and
      Lauscher, Anne",
    editor = "Christodoulopoulos, Christos  and
      Chakraborty, Tanmoy  and
      Rose, Carolyn  and
      Peng, Violet",
    booktitle = "Findings of the Association for Computational Linguistics: EMNLP 2025",
    month = nov,
    year = "2025",
    address = "Suzhou, China",
    publisher = "Association for Computational Linguistics",
    url = "https://aclanthology.org/2025.findings-emnlp.1002/",
    doi = "10.18653/v1/2025.findings-emnlp.1002",
    pages = "18450--18477",
    ISBN = "979-8-89176-335-7"
}

@misc{qi_stanza_2020,
	title = {Stanza: {A} {Python} {Natural} {Language} {Processing} {Toolkit} for {Many} {Human} {Languages}},
	copyright = {arXiv.org perpetual, non-exclusive license},
	shorttitle = {Stanza},
	url = {https://arxiv.org/abs/2003.07082},
	doi = {10.48550/ARXIV.2003.07082},
	urldate = {2026-04-14},
	publisher = {arXiv},
	author = {Qi, Peng and Zhang, Yuhao and Zhang, Yuhui and Bolton, Jason and Manning, Christopher D.},
	year = {2020},
	note = {Version Number: 2},
}
\bibliographystyle{eamt26}

\appendix
\section{Appendix}\label{sec:appendix}

\subsection{Existing Data Resources}\label{app:benchmark_examples}

An overview of existing data resources, including an example sentence each, is provided in Table~\ref{app:tab:dataset_examples}. To highlight a few differences of \textsc{GAND} in relation to existing datasets, we provide specific examples for WinoMT and GATE. While WinoMT also includes one gender-ambiguous referent entity, WinoMT is artificially handcrafted and follows a rigid structure of each sentence referring to two entities, one of which is disambiguated by a pronoun. GATE is a qualitative linguistically diverse corpus of gender-ambiguous source sentences, of which a subset could be considered similar to \textsc{GAND}. It differs from the dataset presented in this paper as it is a ``corpus of hand-curated test cases designed to challenge gender rewriters'' \cite[p.~852]{rarrick-etal-2023}. GATE is partially handcrafted from scratch and partially based on natural, filtered data, of which a subset has been manually modified. GATE covers $\sim$2000 source examples, of which only 1000 include a single referent entity, and 500 include multiple referents whose gender is assessed, and another 500 include single pronoun referents (e.g., ``\textit{I} am tired.''). Their referent entities (which they refer to as ``animate entities'') also include, to a small degree, non-human referents, e.g. `cat'.

\begin{table}[b]
\begin{tabular}{c|c|c}
 & \textbf{OpenSubs}& \textbf{C4} \\
 \midrule
Pre-Auto & 2,750,665 & 20,405,275 \\
\#referent & 19,396 & 1,148,181 \\
Post-CoRef & 9023 & 460,099 \\
\midrule
\midrule
Pre-Manual & 4326 & 10,185 \\
Post-Manual & 2140 & 2908 \\
\midrule
\textsc{GAND} Total & \multicolumn{2}{c}{\textbf{5047}} \\
\bottomrule
\end{tabular}
\caption{\label{app:tab:GAND-stats}
 Number of sentences filtered during the automatic and manual processing steps.
  }
\end{table}

\subsection{\textsc{GAND} Compilation: Extras}\label{GAND_rules}

The list of referent entities and their gender association used to filter data in the creation of \textsc{GAND} is presented in Table~\ref{referent_entities}. The list of gender (pro)nouns used to filter out coreferent sentences is presented in Table~\ref{gender_nouns}. The processing rules for the creation of \textsc{GAND} are included in Table~\ref{coreference}.

\subsubsection{GAND Statistics}\label{app:GAND_stats}
We initially filter 23 million sentences from OpenSubtitles and C4 combined. First, we filter for referent entities, and then we apply the automatic processing filters to exclude coreference. For OpenSubtitles, this leaves 9023 sentences (only 0.33\% of the original), and for C4, this leaves 460,099 (2.25\% of the original). The numbers are outlined in Table \ref{app:tab:GAND-stats}. We select only a handful of these, a total of 14,511, for manual reference to exclude any further sentences that were not excluded in the automatic processing step. This necessary step further excludes another 65.33\%, leaving 5047 for \textsc{GAND}.

\subsubsection{GAND Exclusions}\label{app:exclusions}
Examples of sentences that passed the automatic processing step but were manually excluded from the compilation of \textsc{GAND} to ensure a clean gender-ambiguous dataset referring to a singular referent are included in Table \ref{app:tab:exclusions}. Among others, excluded sentences included wrong entities (e.g., `Acrobat reader'), disambiguating gender cues, sentences with indirect references to the referent, and senseless sentences. The intended referent (for which we filtered) is marked in italics.

\subsection{Probability Difference per Referent}\label{app:prob_diff_referent}
\begin{figure*}
  \includegraphics[width=0.9\textwidth]{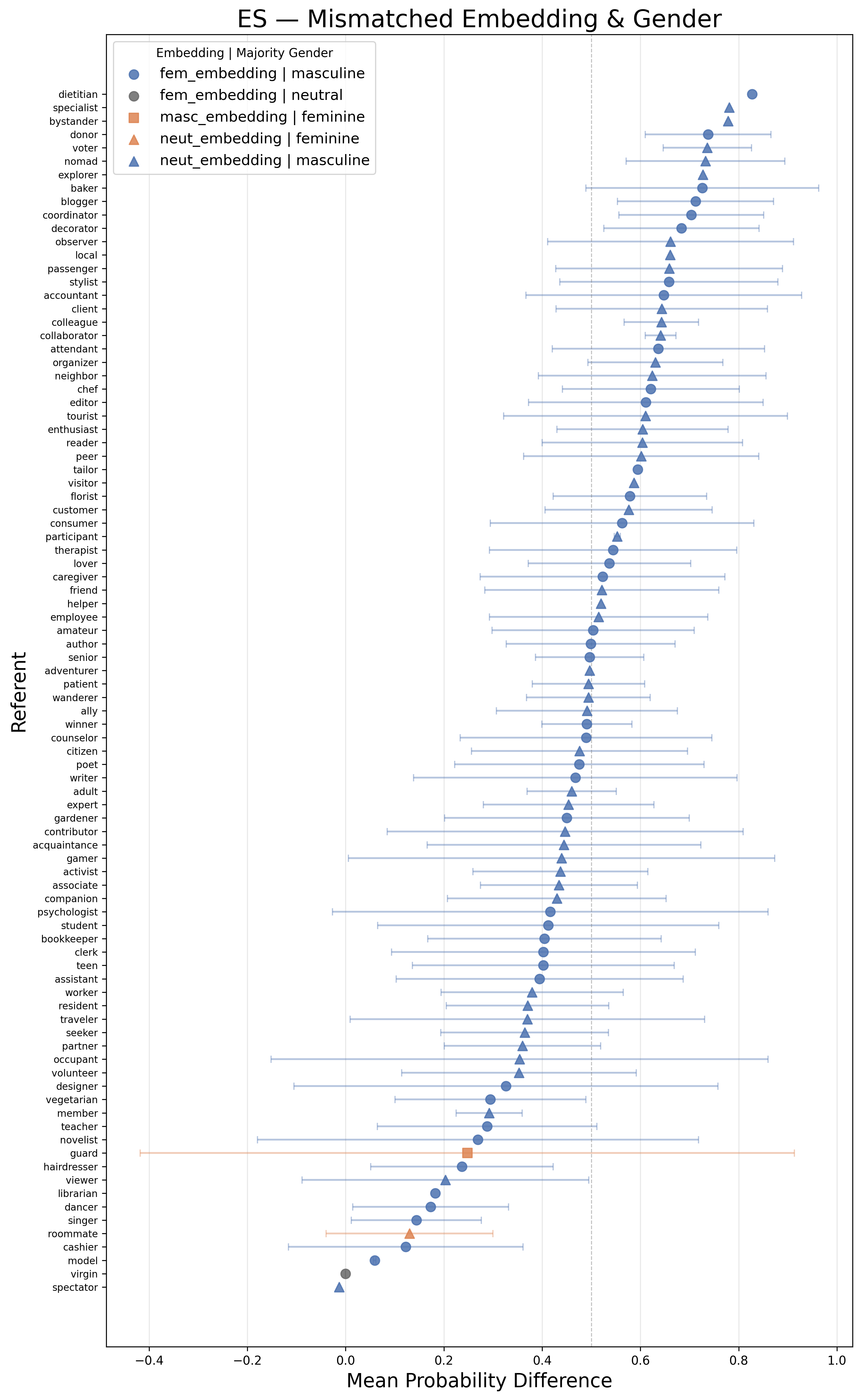}
  \caption{Mean probability difference per referent for Spanish, where there is a mismatch in referent embedding and target gender (e.g., neutral `embedding' but masculine target translation).}
  \label{fig:prob_diff_mismatch_referent_ES}
\end{figure*}

\begin{figure*}
  \includegraphics[width=0.9\textwidth]{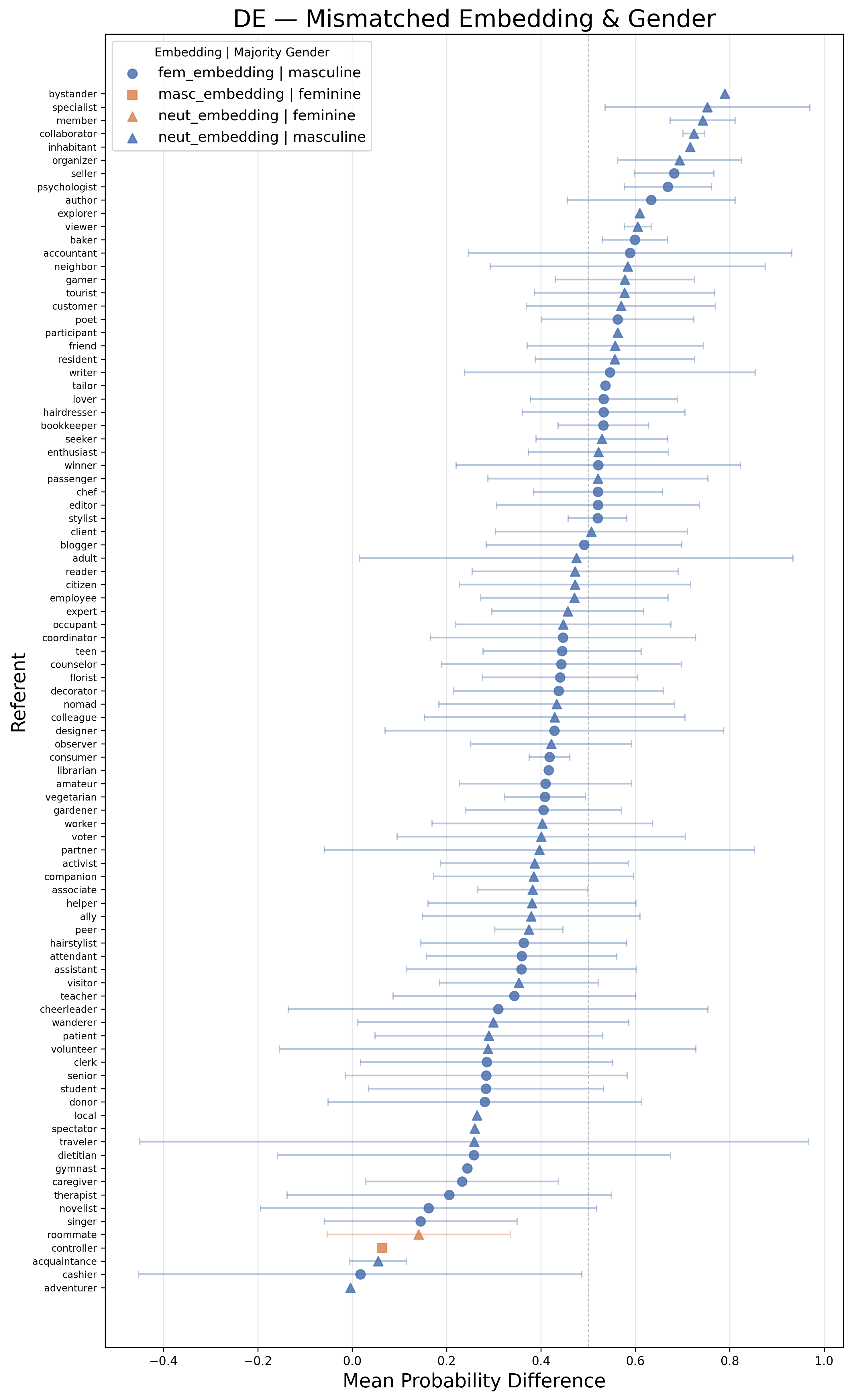}
  \caption{Mean probability difference per referent for German, where there is a mismatch in referent embedding and target gender (e.g., neutral `embedding' but masculine target translation).}
  \label{fig:prob_diff_mismatch_referent_DE}
\end{figure*}

\begin{figure*}
  \includegraphics[width=\textwidth]{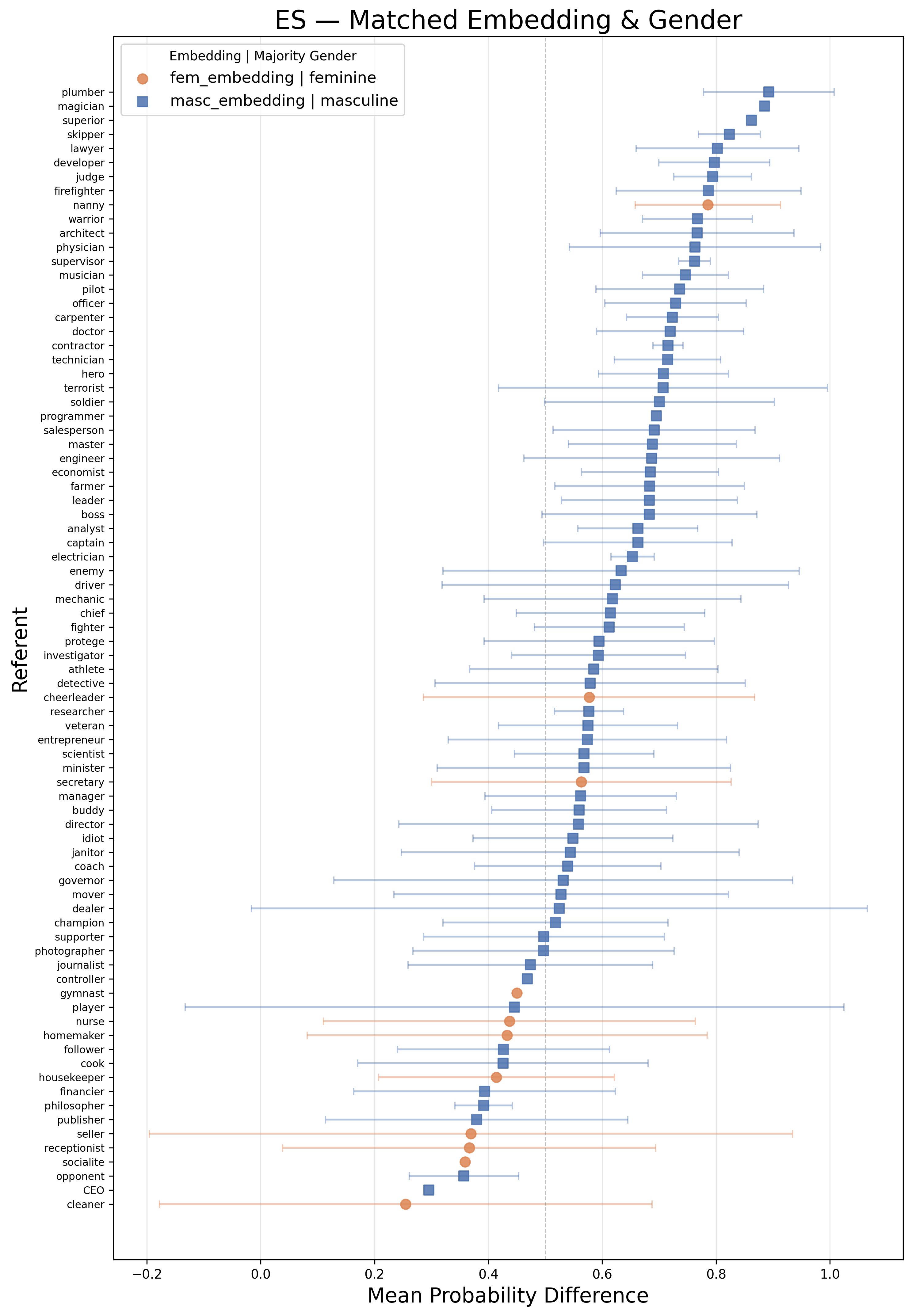}
  \caption{Mean probability difference per referent for Spanish, where there is a match in referent embedding and target gender (e.g., feminine embedding and feminine target translation).}
  \label{fig:prob_diff_match_referent_ES}
\end{figure*}

\begin{figure*}
  \includegraphics[width=\textwidth]{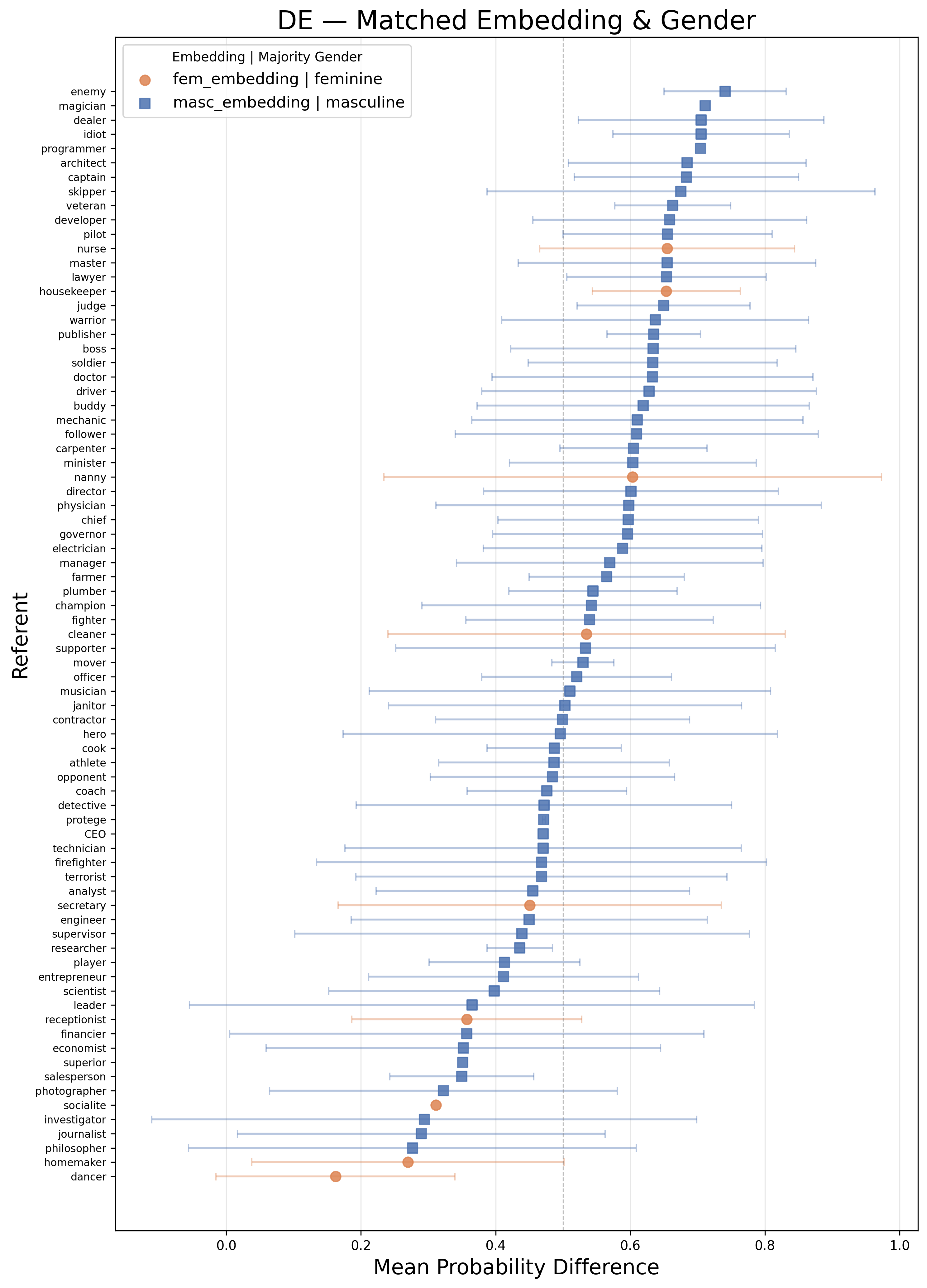}
  \caption{Mean probability difference per referent for German, where there is a match in referent embedding and target gender (e.g., feminine embedding and feminine target translation).}
  \label{fig:prob_diff_match_referent_DE}
\end{figure*}

This subsection presents the probability difference of translating into a certain target gender for each source referent and each language. Figures \ref{fig:prob_diff_mismatch_referent_ES} and \ref{fig:prob_diff_mismatch_referent_DE} show referents (y-axis) where the source embedding is different to the (majority) gender in the target translation, and the mean probability difference of the attribution analysis (x-axis). As an example, the top of Figure \ref{fig:prob_diff_mismatch_referent_ES} shows that the referent `dietitian' has a feminine word embedding but has been translated into masculine (as a majority) with an average probability difference of $\sim$0.8 (i.e. 80\% more likely to be masculine \textit{instead of} feminine). Figures \ref{fig:prob_diff_match_referent_ES} and \ref{fig:prob_diff_match_referent_DE} show referents where the source embedding and the gender in the target translation matched.

\begingroup
\renewcommand{\arraystretch}{1.75} 
\begin{table*}
  \centering
  \begin{tabularx}{\textwidth}{>{\RaggedRight\arraybackslash}p{2.5cm}>{\RaggedRight\arraybackslash}p{4.75cm}X}
    \toprule
    \textbf{Benchmark} & \textbf{Authors} & \textbf{Example Sentence} \\
    \midrule
    WinoBias & Zhao et al. ~\shortcite{zhao-etal-2018-gender}  & ``The \textit{physician} hired the secretary because \textit{she} was overwhelmed with clients.''  \\
    WinoMT & Stanovsky et al. ~\shortcite{stanovsky-etal-2019}  & ``The \textit{doctor} asked the nurse to help \textit{her} in the procedure.'' \\
    SimpleGen & Renduchintala et al. ~\shortcite{renduchintala-etal-2021}  & ``That \textit{physician} is a funny \textit{lady}!'' \\
    BUG & Levy et al. ~\shortcite{levy-etal-2021} & ``With \textit{his} dark hair and complexion, the ballet \textit{dancer} was often cast in more exoitic roles.'' \\
    MT-GenEval & Currey et al. ~\shortcite{currey-etal-2022} & ``Having served \textit{his} apprenticeship Crookall became a master \textit{painter} trading at Duke Street, Douglas.'' \\
    MiTTens & Robinson et al. ~\shortcite{robinson-etal-2024} & ES: ``\textit{Vino} de inmediato cuando se enteró. Es \textit{una buena médica}.'' \\&& EN: ``\textit{He} came immediately when \textit{he} heard about it. \textit{He} is a good doctor.'' \\
    GLITTER & Pranav et al. ~\shortcite{pranav-etal-2025} & ``[...] objectives with enthusiasm. \textit{Colleagues}, such as John and Emily, supported [...]. The culture [...]'' \\
    GeNTE & Piergentili et al. ~\shortcite{piergentili-etal-2023-hi} & ``I, along with all my \textit{colleagues}, wish to welcome this [...].'' \\
    mGeNTE & Savoldi et al. ~\shortcite{savoldi-etal-2025-mgente} & ``EU resources come from European \textit{citizens}.'' \\
    Must-SHE & Bentivogli et al. ~\shortcite{bentivogli-etal-2020} & ``\textit{I} was \textit{born} and \textit{brought up} in Mumbai.'' \\
    Arabic Parallel Gender Corpus & Habash et al. ~\shortcite{habash-etal-2019}  & ``\textit{I} have no interest in that.'' \\
    GATE & Rarrick et al. ~\shortcite{rarrick-etal-2023} & ``A catholic \textit{friend} of mine attended a jewish worship service with me.'' \\&& ``The ideal is six \textit{players} who win and win and a high level.''\\
    GENDEROUS & Hackenbuchner et al. ~\shortcite{hackenbuchner-etal-2025-genderous} & ``The assured \textit{electrician} finished the work.'' \\
    GAMBIT & Menis-Mastromichalakis et al. ~\shortcite{mastromichalakis-etal-2025} & ``The government \textit{minister} announced new policies during yesterday's parliamentary session.'' \\
    GAMBIT+ & Filandrianos et al. ~\shortcite{filandrianos-etal-2025} & ```Did you hear about the press conference yesterday?' `Yeah, the Police \textit{Commissioner} announced new safety measures for the downtown area.'[...]''\\
    
    \bottomrule
  \end{tabularx}
  \caption{\label{app:tab:dataset_examples}
    Each benchmark introduced in Section \ref{related} including one example sentence. This is not an exhaustive list of datasets and many of these presented here contain different types of sentences: a combination of ambiguous and unambiguous, of referent singularity and plurality, and of natural and artificial data etc.
  }
\end{table*}
\endgroup

\begingroup
\renewcommand{\arraystretch}{1.5} 
\begin{table*}
  \centering
  \begin{tabularx}{\textwidth}{lX}
  \toprule
    \textbf{Gender} & \textbf{Lists of Referent Entities} \\
    \midrule
    Female association & homemaker, nurse, receptionist, librarian, socialite, hairdresser, nanny, bookkeeper, stylist, housekeeper, designer, clerk, assistant, teacher, baker, auditor, editor, secretary, cashier, tailor, cleaner, writer, accountant, counselor, attendant, author, blogger, model, vegetarian, dancer, donor, lover, senior, therapist, amateur, teen, consumer, coordinator, seller, student, virgin, winner, caregiver, beautician, dietitian, psychologist, hairstylist, chef, florist, gardener, decorator, influencer, poet, novelist, singer, gymnast, cheerleader\\
    
    Male association & skipper, protege, philosopher, captain, architect, financier, warrior, magician, pilot, boss, developer, mechanic, mover, analyst, chief, salesperson, lawyer, cook, physician, farmer, CEO, manager, driver, guard, laborer, carpenter, janitor, supervisor, champion, user, contractor, engineer, controller, dealer, master, superior, follower, buddy, coach, hero, idiot, musician, player, enemy, fighter, governor, leader, minister, officer, opponent, soldier, supporter, terrorist, veteran, doctor, scientist, programmer, firefighter, technician, plumber, electrician, entrepreneur, director, judge, detective, investigator, researcher, economist, publisher, photographer, athlete, journalist\\
    
    Neither association & adult, citizen, resident, participant, member, passenger, occupant, worker, employee, traveler, visitor, performer, volunteer, patient, customer, client, voter, viewer, spectator, explorer, tourist, reader, gamer, contributor, helper, activist, organizer, specialist, expert, professional, neighbor, colleague, friend, acquaintance, companion, ally, confidant, roommate, classmate, peer, associate, partner, collaborator, inhabitant, local, bystander, observer, nomad, wanderer, adventurer, enthusiast, seeker, vacationer, wayfarer\\
    \midrule
  \end{tabularx}
  \caption{\label{referent_entities}
    List of referent entities and their gender association used to filter data in the creation of \textsc{GAND}.
  }
\end{table*}
\endgroup

\begingroup
\renewcommand{\arraystretch}{1.5} 
\begin{table*}
  \centering
  \begin{tabular}{l}
  \toprule
    \textbf{List of Gender (Pro)Nouns for Coreference Exclusion}\\
    \midrule
    he, she, him, her, his, hers, himself, herself \\
    mother, father, mom, dad, sister, brother, wife, husband, grandmother, grandfather, daughter, son \\
    lady, sir, woman, man, women, men, female, male, girl, boy \\
    \bottomrule
  \end{tabular}
  \caption{\label{gender_nouns}
    List of gender (pro)nouns we use to filter out coreferent sentences in the creation of \textsc{GAND}.
  }
\end{table*}
\endgroup

\begingroup
\renewcommand{\arraystretch}{1.5} 
\begin{table*}
  \centering
  \begin{tabularx}{\textwidth}{lX}
  \toprule
     \multicolumn{2}{l}{\textbf{Automatic Processing Filters to Exclude Coreference}} \\
    \midrule
    \textbf{Rule} & \textbf{Description}\\
    \midrule
    CONJ VERB & Eliminate cases where the sentence does not contain any conjugated verb. \\
    OTHER POS & Eliminate cases where the referent entitiy is not a noun (i.e. person) but another POS. \\
    COMPOUND & Eliminate cases where the referent is part of a compound. \\
    GENDER PROPN CHILDREN & Eliminate cases where the children of the referent include any of gender (pro)nouns or a proper noun in a modyfing function (e.g., ``nmod'' or ``appos''), also allowing for the referent to be in a coordinating construction. \\
    GENDER PROPN CHILDREN HEAD & Eliminate cases where any of the gender (pro)nouns or a proper noun appears as the subject of a verb that is a child of the verbal head of the referent. \\
    GENDER PROPN OBLIQUE & In case the referent is an oblique argument ("obl" dependency relation label), eliminate cases where (1) any of the gender (pro)nouns appears as the child of the (verbal) head and (2) where a proper noun is the subject of the (verbal) head, also allowing the verbal head to be in a coordinating construction. \\
    GENDER PROPN NMOD APPOS & In case the referent is modifying a noun ("nmod" or "appos" dependency relations), eliminate cases where (1) any of the gender (pro)nouns appears as the child of that nominal head or as the subject of the verbal head of that nominal head and (2) where that nominal head is a proper noun (also allowing for the referent to be used in coordination). \\
    GENDER PROPN XCOMP & In case the referent is an open clausal component of a verb ("xcomp" dependency relation), eliminate cases where any of the gender (pro)nouns appears as the child of that verbal head. \\
    GENDER PROPN ROOT & In case the referent is the root of the sentence (which happens with copula verbs), eliminate cases where the subject is any of the gender (pro)nouns or a proper noun (also allowing for the referent to be used in coordination). \\
    GENDER PROPN NSUBJ & In case the referent acts as a subject, eliminate cases where (1) its head is non-verbal and corresponds to either a gender pronoun or a proper noun and (2) its head is verbal and one of the children of that verbal head is a gender pronoun or proper noun in an "xcomp" dependency relation.\\ 
    \hline
  \end{tabularx}
  \caption{\label{coreference}
    Processing rules for the creation of \textsc{GAND} to exclude (co)reference that provides cues that allows the referent to be disambiguated.
  }
\end{table*}
\endgroup

\begingroup
\renewcommand{\arraystretch}{1.5} 
\begin{table*}
\small
  \centering
  \begin{tabularx}{\textwidth}{XX}
    \toprule
    \multicolumn{2}{l}{\textbf{GAND Exclusions}} \\
    \midrule
    \textbf{\makecell[l]{Example sentence}} & \textbf{\makecell[l]{Error/Reason for Exclusion}} \\
    \midrule
    ``How you got your \textit{driver}'s license is beyond me.'' & `Driver's (license)' does not refer to a specific person and would be translated as a compound term (e.g., `Führerschein' (DE) or `licencia de conducir' (ES)).
    \\
    
    ``You will need the Acrobat \textit{reader} in order to view these files.''&`Reader' is not referring to a person. \\
    
    ``I'm a...... fast-food \textit{employee} that cut off his own hand.'' & `His' disambiguates the employee. \\
    
    
    ``Now that salesman has become a respected sales \textit{manager}.''& `Sales\textit{man}' disambiguates the gender of `manager'.\\
    

    ``Your predecessor, Carlos Aribau, was a \textit{friend} of mine.'' & We exclude sentences where the referent could be disambiguated by a name for full ambiguity.\\
    ``They are also full of lessons about love, friendship and life, that will surely leave a lasting impression on the \textit{reader}.'' & Indirect reference to `reader', could be replaced by `a/any reader', or `readers'.\\
    ``Content can be pre-loaded by the company or downloaded by a \textit{user}.'' &Indirect reference to `user, could be replaced by `any user' or `users'.\\
    \bottomrule
  \end{tabularx}
  \caption{\label{app:tab:exclusions}
    Examples of sentences that were manually excluded from the compilation of \textsc{GAND}.
  }
\end{table*}
\endgroup

\newpage

\begin{figure*}
  \includegraphics[width=0.75\textwidth]{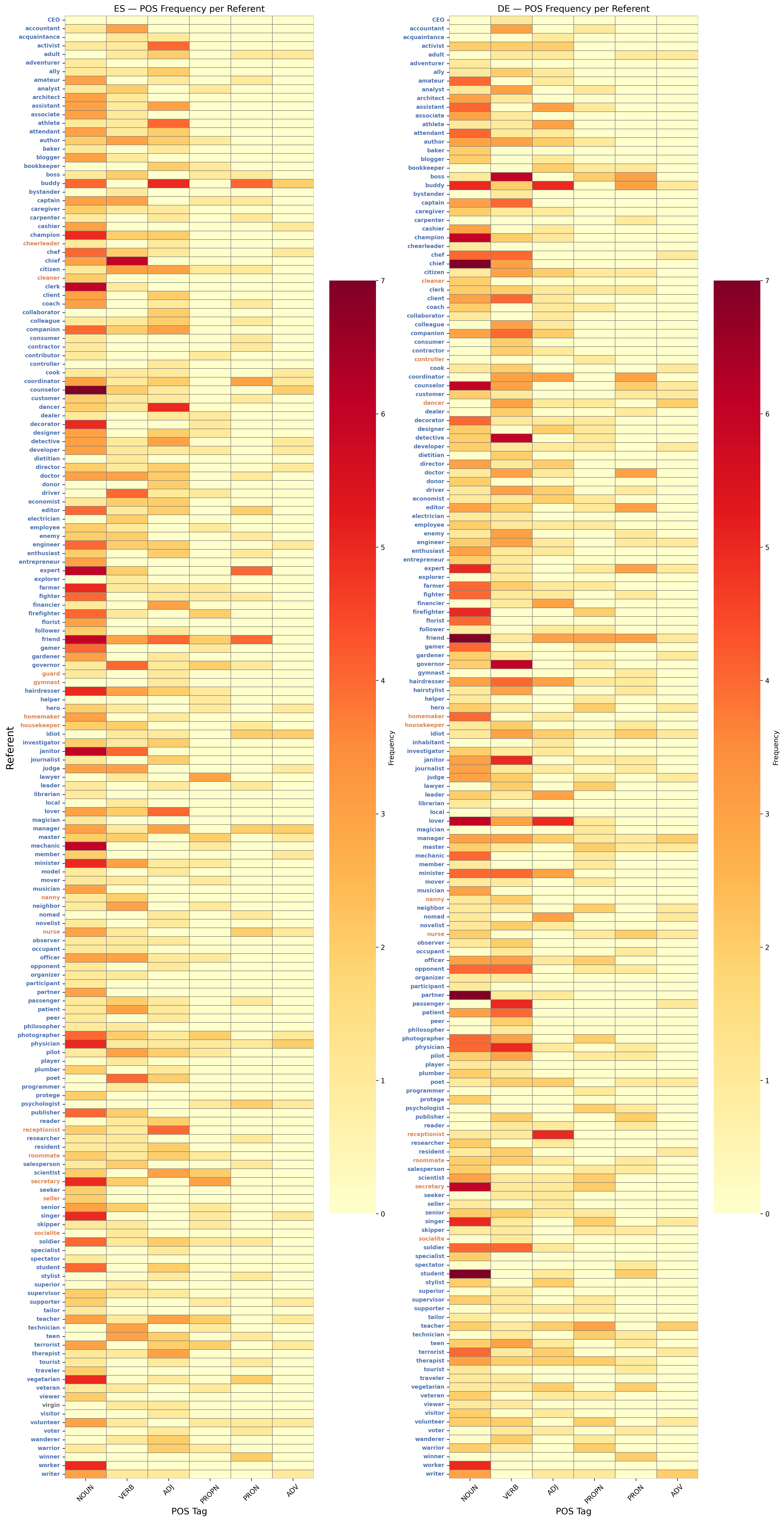}
  \caption{Heatmap of POS tags of salient words for each referent for DE and ES. The x-axis shows POS categories, the y-axis shows (source) referents. Referents that have been translated into masculine are depicted in blue, into feminine are depicted in orange, and into neutral in grey.}
  \label{fig:heatmap}
\end{figure*}

\end{document}